\documentclass[journal]{IEEEtran}

\usepackage[T1]{fontenc}

\usepackage{lineno, hyperref}
\usepackage{nicefrac}
\hypersetup{
colorlinks=true, linktocpage=true, pdfstartpage=3, pdfstartview=FitV,
breaklinks=true, pdfpagemode=UseNone, pageanchor=true, pdfpagemode=UseOutlines,
plainpages=false, bookmarksnumbered, bookmarksopen=true, bookmarksopenlevel=1,
hypertexnames=true, pdfhighlight=/O, urlcolor=blue, linkcolor=blue, citecolor=blue,
    }
\modulolinenumbers[5]

\usepackage{comment}
\usepackage{color}

\usepackage{graphics, dblfloatfix} 
\usepackage{epsfig} 

\usepackage{cite}
\usepackage{amsmath}
\usepackage{amssymb} 
\usepackage{epstopdf, cases}
\usepackage{siunitx}
\usepackage{subfig}	
\usepackage{textcomp}			
\usepackage{booktabs}	
\usepackage{multirow}
\usepackage[absolute]{textpos}	
\usepackage{graphicx}      
\usepackage{tikz}
\usepackage{listings}
\usepackage[normalem]{ulem}
\usepackage{enumitem}
\usepackage{lipsum}
\usepackage[margin=0.9in]{geometry}
\usepackage{siunitx}
\usepackage{multirow}
\usepackage{multicol}
\usepackage{rotating}
\usepackage{tabularray}
\usepackage{pifont}
\usepackage{xcolor}
\usepackage{color, colortbl}
\definecolor{Gray}{gray}{0.9}
\newcolumntype{g}{>{\columncolor{Gray}}c}

\usepackage{algorithm}
\usepackage{algorithmicx}
\usepackage[noend]{algpseudocode}

\makeatletter
\newenvironment{subalgorithm}{
    \renewcommand{\ALG@name}{Sub-Algorithm}
   \begin{algorithm}%
  }{\end{algorithm}}
\makeatother

\newcounter{rootalgorithm}
\newcounter{subalgorithm}

\usetikzlibrary{matrix}

\definecolor{nero}{rgb}{0, 0, 0}
\definecolor{rosso}{rgb}{0.9, 0, 0}
\definecolor{verde}{rgb}{0, 0.6, 0}
\definecolor{blu}{rgb}{0, 0, 0.9}
\definecolor{grigio}{rgb}{0.52, 0.52, 0.51}

\definecolor{revisione}{rgb}{0.0, 0, 0}
 \definecolor{revisione2}{rgb}{0.9, 0, 0}

\newcommand{\rev}[1]{\textcolor{revisione}{#1}}

\newcommand{\revout}[1]{}
\newcommand{\revbar}[1]{}

\makeatletter
\newcommand{\definetrim}[4]{%
 \define@key{Gin}{#1}[]{\setkeys{Gin}{trim=#2, clip, width=#3}}%
}
\makeatother
\definetrim{matlab_figure}{3.5cm 8.5cm 4.6cm 9.2cm}{0.48\textwidth}

\newcommand{\PaperTitle}{Reactive and Safety-Aware Path Replanning for Collaborative Applications}

\title{\PaperTitle}

\author{
Cesare~Tonola$^{1}$, 
Marco~Faroni$^{2}$,
Saeed~Abdolshah$^{3}$,
Mazin~Hamad$^{3}$,\\
Sami~Haddadin$^{3,4}$,
Nicola~Pedrocchi$^{1}$, 
Manuel~Beschi$^{1, 5}$
\thanks{
$^{1}$Institute of Intelligent Industrial Technologies and Systems, National Research Council of Italy, STIIMA-CNR, Milan, Italy. \\\small\tt \{cesare.tonola\}@stiima.cnr.it
}%
\thanks{
$^{2}$Dipartimento di Elettronica, Informazione e Bioingegneria, Politecnico di Milano, Milan, Italy.
}
\thanks{
$^{3}$Munich Institute of Robotics and Machine Intelligence (MIRMI), Technical University Munich, Munich, Germany.
}
\thanks{
$^{4}$Mohamed Bin Zayed University of Artificial Intelligence, Masdar City, Abu Dhabi, UAE.
}%
\thanks{
Munich Institute of Robotics and Machine Intelligence (MIRMI), Technical University Munich, Munich, Germany.
}
\thanks{
$^{5}$Dipartimento di Ingegneria Meccanica e Industriale, University of Brescia, Italy. 
}

}

\begin{document}

\maketitle

\begin{abstract} \label{abstract}
This paper addresses motion replanning in human-robot collaborative scenarios, emphasizing reactivity and safety-compliant efficiency.
While existing human-aware motion planners are effective in structured environments, they often struggle with unpredictable human behavior, leading to safety measures that limit robot performance and throughput.
In this study, we combine reactive path replanning and a safety-aware cost function, allowing the robot to adjust its path to changes in the human state. 
This solution reduces the execution time and the need for trajectory slowdowns without sacrificing safety. 
Simulations and real-world experiments show the method's effectiveness compared to standard human-robot cooperation approaches, with efficiency enhancements of up to 60\%. 
\end{abstract}

\def\abstractname{Note to Practitioners}
\begin{abstract}
The ISO/TS 15066 outlines collaborative operations, such as Speed and Separation Monitoring (SSM), designed to uphold worker safety in human-robot collaboration. While these safety measures lead to robot deceleration to prevent collisions, they can also cause inefficiencies due to robot inactivity. This work introduces a dynamic real-time adaptation of the robot's path, aiming to minimize the execution time while considering safety interventions. In contrast to methods relying on pre-computed trajectories, this approach modifies the robot's path in real-time based on the continuous monitoring of the human state, thereby decreasing safety interventions and downtime, especially in dynamic environments.
\end{abstract}




\section{Introduction} \label{sec: introduction}
\IEEEPARstart{T}{he} factory of the future envisions seamless collaboration between humans and robots as a means to enhance industrial productivity and improve workers' conditions \cite{Robla-Gómez2017}. 
Collaborative applications are subject to safety requirements to mitigate potential hazards to the workers' health \cite{VICENTINI2020101921}.
The ISO/TS 15066 technical specification \cite{ISOTS15066} defines the \textit{Power and Force Limiting} (PFL) and the \textit{Speed and Separation Monitoring} (SSM) modes, which describe safety requirements for applications with and without contacts, respectively.
Typically, their implementation restricts the robot’s speed and halts it when close to a human.

\rev{Traditional motion planners are poorly suited to collaborative settings. They treat humans as obstacles and do not dynamically adapt to human movements. They generally optimize a cost function, such as minimizing path length or travel time,  without considering how safety constraints or human motion may affect the \textit{a-posteriori} execution time. 
Recent research has tackled this issue at different levels, such as task-planning level \cite{SandroSamuelini, faccio2023task}, control level \cite{Huang2016, Dotoli2022}, and motion planning level \cite{10611118,Palleschi2021}.}

\rev{
Previous studies, such as \cite{Palleschi2021, 9143390, BYNER2019239, MARVEL2017144, Tonola-roman, MARS, TPRM}, exemplify \textit{reactive} planning approaches that adapt a robot trajectory in real-time based on human proximity. Works like \cite{Palleschi2021, 9143390, BYNER2019239, MARVEL2017144} modify only the speed and are effective for brief, infrequent intrusions, but they become inefficient with frequent close collaboration, resulting in repeated safety stops when the human is nearby. In contrast, \cite{Tonola-roman, MARS, TPRM} dynamically adjust the robot path and prioritize computational efficiency by searching for feasible, minimum-length solutions but do not consider the impact of safety requirements on the path.}

\rev{
Another class of approaches involves \textit{proactive} planners \cite{Casalino2019, HAMP, Mainprice2011}.
They pre-compute trajectories just before the motion starts by optimizing \textit{ad-hoc} cost functions beyond minimizing path length. For example, they reduce expected interference between the operator and the robot, relying on human behavior models to minimize collision risk~\cite{Casalino2019}, execution time~\cite{HAMP}, or human discomfort~\cite{Mainprice2011}. However, the computational demands of these cost functions generally make them unsuitable for real-time path adjustments.
As a result, proactive planners lack online adaptability, and their effectiveness diminishes in settings where the operator’s movements are unpredictable.}

\rev{In such a context, our goal is to get the best from both reactive and proactive approaches so that we can replan the robot path online according to the human state with a rate compatible with runtime implementation ($\geq 5$ Hz) and minimize the trajectory traversal time during execution.
Our approach uses a safety-aware cost function based on ISO/TS 15066 that continuously estimates the expected trajectory execution time--factoring in potential safety stops and slow-downs--and integrates it into a real-time path replanning algorithm.}
\rev{
To achieve real-time performance, we speed up the trajectory re-computation by (i) restricting the search space through a search heuristic and (ii) exploiting prior search information and employing lazy cost function evaluations. This balances computational complexity with speed, significantly boosting the efficiency of online path search.}
The proposed method is called ``Human-Aware Multi-pAth Replanning Strategy'' (MARSHA), and it allows for continuous adaptation to unexpected human movements while still providing safety-aware solutions. It dynamically estimates the expected execution time of the robot's movements and minimizes it by adjusting the trajectory. We evaluated our method with simulations and real experiments (video available at \cite{video}). A ROS \cite{ROS} compatible algorithm implementation is available at \cite{marsha_examples}.

\section{Related Works} \label{sec: related works}


Human-aware motion planners enable robots to move while considering the human presence. 
Previous works focused on optimizing paths by considering the human-robot distance, the human field of view, and comfort \cite{Sisbot2012, Mainprice2011}. 
Other approaches avoid high-occupancy areas by acting on the planner cost function \cite{Hayne2016, Tarbouriech2020, Zhao2018}. For example, \cite{Hayne2016} biases the sampling to regions situated far from the current human state \cite{Tarbouriech2020}, \cite{Zhao2018} utilized STOMP \cite{STOMP:2011} to minimize a cost function penalizing previously occupied regions while maximizing the human-robot distance, and \cite{Casalino2019} deformed the trajectory based on a repulsion field associated with checkpoints on the human skeleton. 
\cite{9779321} accounts for collision risk, social norms, and crowded areas. 
\cite{Kanazawa2019} focused on reducing execution time and the risk of robot–worker collisions in handover tasks. 
Some studies evaluated human factors and demonstrated enhancements in work fluency and operator satisfaction when a human-aware motion planner is adopted \cite{Przemyslaw2015, BESCHI2020744}.
Recent works dealt with the safety integration within the motion planner. 
\cite{HAMP} and \cite{S*} evaluate paths by estimating the speed scaling when the robot crosses those paths. \cite{flowers2023spatiotemporal} uses spatiotemporal human occupancy maps to find robot trajectories that anticipate human movements.
These methods consider speed limitations in motion generation.
They do not require tuning and minimize an estimate of the execution time according to SSM \cite{HAMP, flowers2023spatiotemporal} and PFL \cite{S*} rules.
However, these methods lack robustness to human behavior variations and real-time adaptability.


Path replanning algorithms allow for quick adjustments to the robot's current trajectory, ensuring fast reactions to environmental changes, \rev{whereas recalculating a path from scratch would be time-consuming and reduce the robot's overall reactivity.}
Four main approaches can be identified for algorithms dealing with real-time path modification:

\subsubsection{Graph-based methods} They build a graph on the discretized search space and modify the path when new obstacle-related information emerges \cite{LPAStar, D*Lite}. Their suitability is limited to small-dimensional problems like mobile robot navigation due to poor scalability.

\subsubsection{Potential fields methods} These methods modify precomputed trajectories using virtual attracting and repelling forces generated by obstacles and goals \cite{37903, Chiriatti21, Liu2022}. Despite the high reactivity, the robot may get stuck in local minima, and it does not guarantee finding a valid solution.

\subsubsection{Learning-based methods} \rev{These methods deliver reactive and efficient performance in environments similar to their training conditions. Most approaches compute the robot’s next action (\textit{e.g.,} position or velocity command) based on the current state of the robot and environment, using techniques such as Reinforcement Learning or Graph Neural Networks \cite{Nicola2021, LUO2023104545, CHEN202264, 10576733}. However, they often struggle to generalize effectively to new environments.}

\subsubsection{Sampling-based methods} These algorithms, like RRT and PRM variants \cite{MARS, TPRM, DRRT, smart, 9844664}, are commonly used for path planning without upfront discretization. They are popular thanks to their real-time adaptability, scalability, and robustness to dynamic environments and complex scenarios.
Recently, a sampling-based Multi-pAth Replanning Strategy (MARS) \cite{MARS} has been proposed by the authors to address replanning in complex search spaces. MARS uses a set of precomputed paths to find a solution and refine it over time rapidly. 
The algorithm reuses existing subtrees and precomputed paths during the optimization and it exploits a parallel architecture.
%
 However, it focuses on finding the shortest path, without embedding any effects the safety module may have on the path itself, leading to further slowdowns.

\section{Preliminaries} \label{sec: preliminaries}
\subsection{Path Planning Formulation}
Path planning involves determining a path $\sigma$ (\textit{i.e.}, a sequence of feasible robot's configurations) that connects an initial configuration $q_{\text{start}} \in \mathcal{C}$ to a final configuration $q_{\text{goal}} \in \mathcal{C}$. Here, $\mathcal{C} \subseteq \mathbb{R}^n$ is the robot's configuration space, $n$ is the number of degrees of freedom of the robot, and $q$ is the vector of robot's joint angles. A configuration $q$ is considered feasible if it belongs to the collision-free configuration set $\mathcal{C}_{\text{free}} \subseteq \mathcal{C}$. 
A cost function assigns a non-negative scalar cost $c$ to the path. 
The optimal path $\sigma^*: [0, 1] \rightarrow \mathcal{C}_{\text{free}}$ is the one that minimizes the cost function:
\begin{equation}
\sigma^* = \arg\min_{\sigma} c(\sigma) \quad \text{s.t.} \quad \sigma(0) = q_{\text{start}}, \sigma(1) = q_{\text{goal}},
\end{equation}
A common choice for $c$ is the path length. For example, if $\sigma$ is discretized into $w$ waypoints, we have:

\begin{equation}
c(\sigma) = \sum_{i=1}^{w-1} ||q_{i+1} - q_{i}||_2
\end{equation}
In the context of safety-aware path planning, \cite{HAMP} evaluates the cost as the estimated path execution time $t_{\text{est}}$ which embeds safety constraints based on the human state $\mathcal{H}$:
\begin{equation} \label{eq: hamp_cost_fcn}
c(\sigma) = t_{\text{est}}(\sigma, \mathcal{H})
\end{equation}
By discretizing $\sigma$, \eqref{eq: hamp_cost_fcn} can be expressed as follows:
\begin{equation} \label{eq: hamp_cost_fcn2}
c(\sigma) = \sum_{i=1}^{w-1} t_{\text{nom}, i} \lambda_{i} \left(q_{i}, q_{i+1}, \mathcal{H}\right)
\end{equation}
where $t_{\text{nom}, i}$ is the nominal execution time (no human in the scene), and $\lambda_{i}$ is the average time dilation factor (see Section \ref{sec: cost function}) that measures the effect of the human on the execution time of connection $\overline{q_{i}q_{i+1}}$, \textit{i.e.}:

\begin{equation} \label{eq: lambda}
\lambda_i = \frac{t_{\text{est}, i}(\mathcal{H})}{t_{\text{nom}, i}} \geq 1
\end{equation}
Note that $t_{\text{nom}, i}$ is usually obtained \textit{a posteriori} using a path parametrization method (\textit{e.g.}, TOPP \cite{TOPP}) as each segment's timing is influenced by the kinematic constraints of adjacent segments. It can be underestimated by considering the minimum time required to cross the connection $\overline{q_{i}q_{i+1}}$, imposing that at least one joint is moving at its maximum speed. This leads to an overestimation of the velocity in favor of safety. The minimum time required to travel $\overline{q_{i}q_{i+1}}$ is therefore given by the maximum of the component-wise ratio $(q_{i+1} - q_{i}) \oslash \dot{q}_\text{max}$, where $\dot{q}_\text{max}$ denotes the maximum joint speed vector; summarizing, \eqref{eq: hamp_cost_fcn2} becomes:
\begin{equation} \label{eq: hamp_cost_fcn3}
c(\sigma) = \sum_{i=1}^{w-1} \Big\| (q_{i+1} - q_{i}) \oslash \dot{q}_\text{max}\Big\|_\infty \lambda_i \left(q_{i}, q_{i+1}, \mathcal{H}\right)
\end{equation}

\subsection{Safety Requirements} \label{sec: safety}
Safety requirements for collaborative applications are defined by ISO/TS 15066 (\textit{Robots and robotic devices – Collaborative robots} \cite{ISOTS15066}). 
The document outlines various collaborative operations, with two of particular interest.

\subsubsection{Speed and Separation Monitoring (SSM)} focuses on maintaining a minimum distance between the human and the robot to ensure the robot can stop before possible impacts. Specifically, the minimum human-robot distance, denoted as $S$, should always satisfy the condition 
$   S \geq S_p$.
Here, $S_p$  represents the protective separation distance. 
We can derive the maximum safe velocity $v_{max}$ of the robot towards the human based on the current separation distance $S$:
\begin{equation} \label{eq: vmax SSM}
    v_{max} = \sqrt{v_h^2 + (a_s T_r)^2 - 2 a_s \big(C-S\big)} - a_s T_r - v_h
\end{equation}
where $v_h$ is the human velocity towards the robot, $a_s$ is the maximum Cartesian deceleration of the robot towards the human, $T_r$ is the reaction time of the system, and $C$ is a parameter accounting for the uncertainty of the perception system.
\subsubsection{Power and Force Limiting (PFL)} allows the robot to come into contact with a human with non-zero speed, as long as the kinetic energy transferred to the human does not exceed a threshold $E_{max}$. $E_{max}$ is turned into a limit on the maximum robot speed:
\begin{equation} \label{eq: vmax PFL}
    v_{max} = \frac{F_{max}}{\sqrt{k}} \sqrt{m_{r} ^{-1} + m_{h} ^ {-1}}
\end{equation}
where $F_{max}$ and $k$ are the maximum contact force and spring constant for a specific body region and $m_r$ and $m_h$ are the effective mass of the robot and of the human body region, respectively. See \cite{ISOTS15066, Haddadin2012} for more information.


Both SSM and PFL constrain the robot's maximum speed. For brevity, we will focus exclusively on SSM in the experimental campaign described in Sections  \ref{sec: qualitative_assessment} and \ref{sec: case_study}, without compromising the generality of our methodology.

\section{Proposed approach} \label{sec: proposed_approach}
\begin{figure}[t]
    \centering
    \includegraphics[width=0.9\columnwidth]{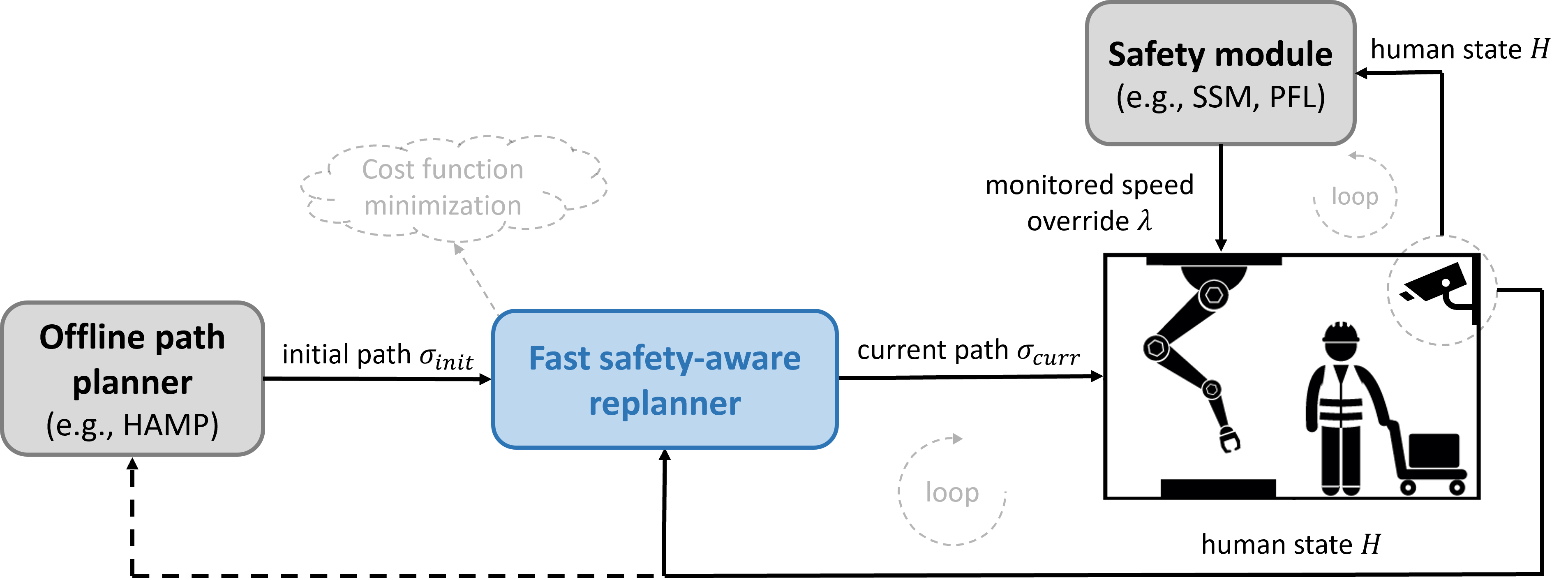}
    \caption{System overview.}
    \label{fig: pipeline}
\end{figure}

The proposed approach operates within the framework of Fig. \ref{fig: pipeline}. 
This paradigm consists of an \textit{offline path planner} responsible for calculating the initial path and a \textit{fast safety-aware replanner}, online updating the trajectory based on information about the human state. 
Additionally, the offline planner can proactively compute a trajectory that reduces the activation of safety rules. 
A safety module implements ISO/TS 15066 to ensure safe cooperation. 
The replanner modifies the path aiming to minimize the intervention required by the safety module, thereby enhancing the overall efficiency of cooperation. 
This goal is achieved by minimizing a safety-aware cost function derived from \eqref{eq: hamp_cost_fcn3} and providing an \textit{admissible informed set} for the cost function.
Including an admissible informed set in a sampling-based path planner speeds up the convergence rate by discarding areas that do not contain the optimal solution, accelerating the online replanning process. 

\subsection{The Safety-Aware Cost Function} \label{sec: cost function}
The main objective of this cost function is to minimize the execution time while considering the human state. 
Simultaneously, in Section \ref{sec: admissible_informed_set}, we define an admissible informed set that can be directly sampled to accelerate the replanning process. The proposed cost function is:
\begin{equation} \label{eq: our_cost_fcn}
c(\sigma) = \sum_{i=1}^{w-1} \left\| (q_{i+1} - q_{i}) \oslash \dot{q}_\text{max}\right\|_2\lambda_i \left(q_{i}, q_{i+1}, \mathcal{H}\right)
\end{equation}
The cost function calculates the length of the path segments, weighting each joint by its maximum speed. Additionally, it incorporates the penalty term $\lambda$ to discourage paths that reduce collaboration efficiency. 
It is worth noting that  \eqref{eq: our_cost_fcn} serves as an upper limit compared to  \eqref{eq: hamp_cost_fcn3}. 
In fact, \eqref{eq: our_cost_fcn} uses the $\mathcal{L}_2$ norm instead of the $\mathcal{L}_\infty$ norm and  $\| v \|_2 \geq \| v \|_\infty$, $\forall v \in \mathbb{R}^{n}$.
Thus, we have:
\begin{equation}
\begin{split}
\sum_{i=1}^{w-1} \| (q_{i+1} - q_{i}) \oslash \dot{q}_\text{max}\|_2\lambda_i \left(q_{i}, q_{i+1}, \mathcal{H}\right) \geq \\ \sum_{i=1}^{w-1} \| (q_{i+1} - q_{i}) \oslash \dot{q}_\text{max}\|_\infty \lambda_i \left(q_{i}, q_{i+1}, \mathcal{H}\right)
\end{split}
\end{equation}
Therefore, by minimizing the proposed cost function, we decrease the upper bound of  \eqref{eq: hamp_cost_fcn3} and consequently reduce the estimated execution time of the path.

The safety guidelines outlined in ISO/TS 15066 can be translated into a human-robot relative speed limit, imposing that the robot's velocity towards the human must not exceed the maximum allowed robot speed  $v_{max}$ according to Section \ref{sec: safety}.
If the nominal robot velocity towards the human $v_{rh}$ is greater than $v_{max}$, the robot will be slowed down by a factor $\nicefrac{v_{rh}}{v_{max}} > 1$, resulting in an expected time dilation factor $\lambda > 1$.
%
%
We can compute \(v_{rh}\) and $\lambda$ for a generic configuration $q$ according to \cite{HAMP} as follows.
We consider a set of $m$ points of interest for the human state $\mathcal{H} = \{h_1, h_2, \dots, h_m\}$ and a set of $p$ points of interest on the robot's structure $\mathcal{R} = \{r_1, r_2, \dots, r_p \}$.
\rev{Each tuple $\langle r_j, h_k \rangle \in \mathbb{R}^3 \times \mathbb{R}^3$ denotes the Cartesian positions of a pair of key points, with one point on the robot structure and the other on the human skeleton.}
Each point $r_j$ is associated with the robot's configuration $q$ through the forward kinematic $r_j = \text{fk}_j(q)$. Moreover, $u_{\langle r_j, h_k \rangle}$ is the unit vector from $r_j$ to $h_k$:
\begin{equation}
    \rev{u_{\langle r_j, h_k \rangle}=\frac{h_k-r_j}{\| h_k-r_j\|_2}}
\end{equation}
and $\dot{r}_j$, $\dot{h}_k$ are the velocities of $r_j$ and $h_k$, respectively. \rev{The scalar velocity $v_{\langle r_j, h_k \rangle}$ of the robot point $r_j$ towards the human point $h_k$ is computed as:}
\begin{equation}
        v_{\langle r_j, h_k \rangle} = (\dot{r}_j-\dot{h}_k) \cdot u_{\langle r_j, h_k \rangle} = (J_j(q)\dot{q}-\dot{h}_k) \cdot u_{\langle r_j, h_k \rangle}
\end{equation}
\rev{where the operator~$\cdot$ denotes the scalar (dot) product between two vectors and $J_j(q)$ denotes the Jacobian of $r_j$.
A positive value of $v_{\langle r_j, h_k \rangle}$ is used when the distance between $r_j$ and $h_k$ is decreasing.}\footnote{\rev{The sign of $v_{\langle r_j, h_k \rangle}$ depends on the direction of $u_{\langle r_j, h_k \rangle}$ (from robot to human) and the velocity $(\dot{r}_j-\dot{h}_k)$ (from human to robot).}}
As stated in Section \ref{sec: preliminaries}, robot velocity is overestimated in favor of safety by considering that at least one joint is moving at its maximum speed. Hence, for connection $\overline{q_{i}q_{i+1}}$ we have:

\begin{equation} \label{eq: qp_i}
\begin{aligned}
  u_i &= \frac{(q_{i+1}-q_i)}{||q_{i+1}-q_i||_2}, & \dot{q}_i &=  \Big (\min_{l = 1, \dots, n} \Big |{\frac{\dot{q}_{\text{max}, l}}{u_{i, l}}} \Big | \Big ) u_i
\end{aligned}
\end{equation}
where $n$ is the robot's degrees of freedom, and $l$ is the joint index. 
In \eqref{eq: lambda} and \eqref{eq: our_cost_fcn}, $\lambda_i$ represents the estimated time dilation factor experienced by the robot while traversing the connection $\overline{q_{i}q_{i+1}}$. However, since we are assuming a constant robot velocity along the connection, it can also be computed as the ratio between velocities $v_{rh}$ and $v_{{max}}$.
Thus, the time dilation factor $\lambda(q)$ at a given robot configuration $q$ can be calculated as:
\begin{equation} \label{eq: v_rh/v_max}
    \lambda (q) = \max_{\substack{j = 1, \dots, p \\ k = 1, \dots, m}} {\Big (\frac{v_{\langle r_j, h_k \rangle}}{v_{max}}, 1\Big)} \geq 1
\end{equation}
%
where both $v_{\langle r_j, h_k \rangle}$ and $v_{{max}}$ are scalar quantities.
Equation \eqref{eq: v_rh/v_max} computes $\lambda(q)$ for a single configuration $q$. To determine $\lambda_i$ for the entire connection, we divide $\overline{q_{i}q_{i+1}}$ into $z$ equally spaced configurations $q_\zeta$, with $\zeta = 1, \dots, z$, compute the slowdown factor for each configuration $\lambda_{\zeta} = \lambda(q_\zeta)$, and finally calculate the average slowdown as $\lambda_i = \frac{1}{z} \sum_{\zeta=1}^z \lambda_\zeta$.
The slowdown computation can be parallelized for efficiency. 
By substituting  \eqref{eq: vmax SSM} or \eqref{eq: vmax PFL} in  \eqref{eq: v_rh/v_max} makes the cost function \eqref{eq: our_cost_fcn} penalizing the paths that fail to meet the SSM or PFL requirements.
\subsection{An Admissible Informed Set for the Cost Function} \label{sec: admissible_informed_set}

Given a positive cost function $c(\cdot)$, the cost of the optimal path from $q_\text{start}$ to $q_\text{goal}$ passing through a generic point $q \in \mathcal{C}_\text{free}$ is equal to the optimal cost-to-come (from $q_\text{start}$ to $q$), plus the optimal cost-to-go (from $q$ to $q_\text{goal}$). 
If a heuristic function that never overestimates the cost function is available, it is possible to discard regions of the sampling space that will never contain the optimal path, thus speeding up the planning process.
Such function is called an \emph{admissible heuristic} and the remaining sampling space is known as an \emph{admissible informed set}. 

For instance, if $c(\cdot)$ is the path length, the Euclidean norm is an admissible heuristic because:
\begin{equation} \label{eq: heuristic}
\|q - q_\text{start}\|_2 + \|q_\text{goal} - q\|_2 \leq \min \big(c(q_\text{start},q)  + c(q,q_\text{goal}) \big)
\end{equation}
where the right-hand side represents the cost of the optimal path from $q_\text{start}$ to $q_\text{goal}$ passing through $q$.

Let $ c_{\text{best},i}$ be the cost of best solution found up to iteration $i$ of the planner. 
Because $\min \big(c(q_\text{start},q)  + c(q,q_\text{goal}) \big) \leq  c_{\text{best},i}$, it follows that:
\begin{equation}
\label{eq: ellipse}
    \|q - q_\text{start}\|_2 + \|q_\text{goal} - q\|_2  \leq c_{\text{best},i}
\end{equation}
Suppose \eqref{eq: ellipse} does not hold for a point $q$.
Then, the best possible path through $q$ has a higher cost than $c_{\text{best},i}$, meaning that $q$ cannot belong to the optimal path. 
In general, all points outside the admissible informed set
$
{\mathcal{I}} = \{q \in \mathcal{C}_{\text{free}} \mid || q-q_{\text{start}} ||_2 + || q_{\text{goal}} - q ||_2 < c_{\text{best},i} \}
$
are guaranteed not to improve $c_{\text{best},i}$.
By randomly sampling directly from ${\mathcal{I}}$, rather than from the entire configuration space, we speed up the planner convergence rate  \cite{Gammel-InformedRRT-IROS}. 

Considering that $\lambda\geq 1$ in \eqref{eq: our_cost_fcn}, it is possible to define an admissible heuristic for \eqref{eq: our_cost_fcn} as:
\begin{equation} \label{eq: our_heuristic}
\tilde{c}(q) = \| (q - q_\text{start}) \oslash \dot{q}_\text{max}\|_2 + \| (q_\text{goal} - q) \oslash \dot{q}_\text{max}\|_2
\end{equation}
from which the following admissible informed set derives:
\begin{equation} \label{eq: our_ellipse}
\begin{split}
\tilde{\mathcal{I}} = \{q \in \mathcal{C}_{\text{free}} \mid || (q-q_{\text{start}})\oslash \dot{q}_\text{max} ||_2 \\ + || (q_{\text{goal}} - q)\oslash \dot{q}_\text{max} ||_2 & < c_{\text{best},i}  \}
\end{split}
\end{equation}
Note that  \eqref{eq: our_heuristic} differs from  \eqref{eq: heuristic} because  \eqref{eq: our_heuristic} weighs each joint by the inverse of its maximum speed. 
If we define $\hat{q} = \hat{g}(q) = q \oslash \dot{q}_\text{max}$ and $\hat{\mathcal{C}}_\text{free} = \hat{g}(\mathcal{C}_\text{free})$, \eqref{eq: our_ellipse} becomes:
\begin{equation} \label{eq: our_ellipse2}
\hat{\mathcal{I}} = \{\hat{q} \in \hat{\mathcal{C}}_{\text{free}} \mid || \hat{q}-\hat{q}_{\text{start}} ||_2 + || \hat{q}_{\text{goal}} - \hat{q} ||_2 < c_{\text{best},i} \}
\end{equation}
Hence, it is possible to sample from the informed set \eqref{eq: our_ellipse} by directly sampling from \eqref{eq: our_ellipse2} (see \cite{Gammel-InformedRRT-IROS} for efficient sampling algorithms). 
The sampled point $\hat{q}$ is then mapped back to the original space using the inverse transformation $q = \hat{g}^{-1}(\hat{q})$, yielding a valid sample $q$ from \eqref{eq: our_ellipse}.

\subsection{The Path Replanning Algorithm} \label{sec: replanning}

\renewcommand{\algorithmicrequire}{\textbf{Input:}}
\algrenewcommand\algorithmicloop{\textbf{Thread}}
\algrenewcommand\algorithmicindent{1em}%
\begin{algorithm*}
\caption{The threads that operate in parallel to run and support MARSHA} \label{alg: threads}
\footnotesize
\begin{algorithmic}[1]
\vspace{-16pt}
\begin{multicols}{3}

\Loop\; \emph{Trajectory execution} \;
\State $t = t+\Delta t$
\State $state \leftarrow \mathrm{sampleTrajectory}(\tau, t)$ \; \label{line: sample}
\State $\mathrm{sendToController}(state)$ \; \label{line: send}
\EndLoop
\columnbreak
\Loop\; \emph{Collision check} \;
\State \textcolor{blu}{$\mathcal{H} \leftarrow \mathrm{updateHumanState()}$} \label{line: human_state}
\State $\sigma_{\mathrm{subpath}} \leftarrow \sigma_{\mathrm{curr}}[q_{\mathrm{curr}}, q_{\mathrm{goal}}]$ \label{line: subpath}
\State $\mathcal{P} \leftarrow \mathcal{S} \cup \sigma_{\mathrm{subpath}}$; \label{line: P}
\For {${\sigma_j \in \mathcal{P}}$} \label{line: for_check}
 \State $free \leftarrow \mathrm{checkCollision}(\sigma_j)$\label{line: check}
 \If {$free$}
 \State \textcolor{blu}{$c_{\sigma_j} \leftarrow c(\sigma_j, \mathcal{H})$}\; \label{line: cost}
 \Else
 \State $c_{\sigma_j} \leftarrow +\infty$\;
 \EndIf
\EndFor \label{line: end_for_check}

\EndLoop
\columnbreak
\Loop\; \emph{Path replanning}
\State $q_{\mathrm{curr}} \leftarrow \mathrm{projectOnPath}(state, \sigma_{\mathrm{curr}})$\; \label{line: projectOnPath}
\State \textcolor{blu}{$\sigma_{\mathrm{RP}}, solved \leftarrow$
 \Statex \qquad\qquad $\mathrm{replan}(\sigma_{\mathrm{curr}}, q_{\mathrm{curr}}, \mathcal{S}, \mathrm{\Delta t_{max}}, \mathcal{H})$}\label{line: replan}
\If {$solved$}
 \State $\sigma_{\mathrm{curr}} \leftarrow \sigma_{\mathrm{RP}}$\; \label{line: new_path}
 \State $\tau \leftarrow \mathrm{computeTrajectory}(\sigma_{\mathrm{curr}})$ \; \label{line: new_trj}

\EndIf
\EndLoop
\end{multicols}
\vspace{-12pt}
\end{algorithmic}
\end{algorithm*}


\algnewcommand\algorithmicforeach{\textbf{for each}}
\algdef{S}[FOR]{ForEach}[1]{\algorithmicforeach\ #1\ \algorithmicdo}
\renewcommand{\algorithmicrequire}{\textbf{Input:}}

\begin{algorithm}
\caption{MARSHA: high-level description}\label{alg: high-level description}
\footnotesize
\begin{algorithmic}[1]

\algrenewcommand\algorithmicindent{1em}%

\State Define $\mathcal{P}$ as the set of available paths
\State Define $\mathcal{Q}_1$ as the queue of the nodes of $\sigma_{\mathrm{curr}}$ between $q_{\mathrm{curr}}$ and the obstacle
\State Sort $\mathcal{Q}_1$ by some criterion
\ForEach {node $q_n \in \mathcal{Q}_1$}
    \State Insert the nodes of each $\sigma \in \mathcal{P}$ into a queue $\mathcal{Q}_2$
    \State Sort $\mathcal{Q}_2$ by some criterion
    \ForEach {$q_j \in \mathcal{Q}_2$} \label{line: pathSwitch_section}
        \State Define the informed set on the best cost up to now
        \State Get the subtree rooted in $q_n$ from the informed set
        \State \textcolor{blu}{Grow the subtree in the informed set to reach $q_j$ with lazy cost}
        \Statex \qquad  \textcolor{blu}{evaluation}  \label{line: red1}
    \EndFor \label{line: end_pathSwitch_section}
    \State Update $\mathcal{Q}_1$ if a solution was found
\EndFor
\State Get the best path from the graph

\end{algorithmic}

\end{algorithm}


\begin{subalgorithm}[t]
\caption{Connect Node to Paths}\label{alg:pathswitch}
\footnotesize
\begin{algorithmic}[1]
\algrenewcommand\algorithmicindent{1em}%
\color{revisione}
\Function{growInEllipsoid}{$\mathcal{T}$, $E_{ll}$, $q_j$, $t_{\mathrm{MAX}}$}
\While{$\neg$ $ok$ $\&$ $iter< \mathrm{max\_iter}$ $\&$ $t<t_{\mathrm{MAX}}$}
 \State $iter = iter+1$
 \State $q \leftarrow \mathrm{sampleInEllipsoid}(E_{ll})$
 \State \textcolor{blu}{$q_{\mathrm{new}} \leftarrow \mathcal{T}.\mathrm{growTree}(q)$ \algorithmiccomment{no cost function evaluation} }\label{line:growTree}
 \If{$||q_{\mathrm{new}} - q_j|| < \mathrm{min\_dist}$}
 \If{path from tree root to $q_j$ is valid} \label{alg: start_lazy}
 \State \textcolor{blu}{compute the cost function for the path found}\label{line:cost_eval}
 \State create a second-order connection from $q_{\mathrm{new}}$ to $q_j$\label{line:2orderconn}
 \State $ok\leftarrow \mathrm{True}$
 \Else
 \State hide the invalid branch temporarily from $\mathcal{T}$
 \EndIf \label{alg: end_lazy}
 \EndIf
\EndWhile
\State \Return{$ok$}
\EndFunction

\end{algorithmic}
\end{subalgorithm}

\color{black}


\rev{In this section, we present the Human-Aware Multi-pAth Replanning Strategy (MARSHA), which extends MARS \cite{MARS} to handle the new cost function while maintaining responsiveness and replanning capabilities.}

MARS is a sampling-based anytime path replanning algorithm that operates on a set $\mathcal{S}$ of pre-calculated paths connecting the same start-goal pair. MARS connects the current path to nodes on the available paths rather than directly connecting to the goal and uses the available subpath from those nodes toward the goal. Once an initial solution is obtained, MARS employs heuristics and informed sampling to refine the solution over time iteratively. The algorithm reuses valid subtrees in connecting the current path with the available nodes to reduce computational overhead.
It is worth noting that MARS has been proven to be \textit{probabilistic complete} and \textit{asymptotic optimal} when used in conjunction with an asymptotically optimal planner for subtree growth (see \cite{MARS} for details).

The algorithm uses three parallel threads to maximize performance.
The \textit{Trajectory execution} thread interpolates the trajectory at a high rate and sends corresponding commands to the robot's controller.
The \textit{Path replanning} thread handles the replanning process, continually looking for new paths, refining the solution, and computing a trajectory based on it.
The \textit{Collision check} thread checks the collisions along the current path $\sigma_{\mathrm{curr}}$ and the available paths $\mathcal{S}$.
{\rev{Algorithms \ref{alg: threads}, \ref{alg: high-level description}, and \ref{alg:pathswitch} show the algorithmic changes to obtain MARSHA from MARS.} 
%
%
We adopted a lazy approach that minimizes the need for frequent evaluations of the cost function (line \ref{line: red1} of Algorithm \ref{alg: high-level description}) to tackle the computational load caused by the cost function, which is significantly more demanding than the simple Euclidean norm used in MARS. 
During the subtree expansion, our algorithm employs a strategy that only evaluates the cost of new connections when necessary. 
If the selected path planning algorithm does not use a rewiring procedure (\textit{e.g.}, RRT \cite{RRT}), the cost of connections is assessed after the subtree has successfully connected to its target node (lines \ref{line:growTree} and \ref{line:cost_eval} of Algorithm \ref{alg:pathswitch}), limiting the cost function calculation to the connections that belong to the solution.

The evaluation of the cost function requires the updated human state, $\mathcal{H}$ (line \ref{line: replan} of Algorithm \ref{alg: threads}). 
For example, a human tracking system can provide $\mathcal{H}$.
The collision check thread also updates $\mathcal{H}$ (line \ref{line: human_state}) and calculates the cost of the connections for the paths accordingly (line \ref{line: cost}). This approach ensures that MARSHA does not waste time evaluating the cost of connections to the current path but focuses on the connections belonging to the new solutions that have not yet been evaluated in the iteration. 
For further details on sub-algorithms that remain unchanged from MARS, readers may refer to \cite{MARS}.


\section{Scenario analysis} \label{sec: qualitative_assessment}
\begin{table*}[]
\centering
\begin{tabular}{|l|c|c|c|c|}
\hline
\rowcolor{Gray} 
 & reactivity & human-aware & safety-aware & robustness to changing \\
\rowcolor{Gray} 
 &  & proactivity & proactivity &  working conditions  \\ \hline

\cellcolor{Gray} Standard planners \cite{RRT*, Gammell:BIT, 9037111} &  &  & & \ding{51}    \\ \hline
\cellcolor{Gray} Human-aware planners \cite{Casalino2019, Mainprice2011, Hayne2016} &  & \ding{51} & & \ding{51}  \\ \hline
\cellcolor{Gray} Safety-aware planners \cite{HAMP, S*, flowers2023spatiotemporal} &  & \ding{51} & \ding{51} & \ding{51}\\ \hline
\cellcolor{Gray} Reactive replanners \cite{MARS, TPRM, DRRT, smart} & \ding{51}  &  &  & \ding{51}  \\ \hline
\cellcolor{Gray} Learning-based planners \cite{Nicola2021, LUO2023104545, CHEN202264, 10576733} & \ding{51} & \ding{51}* &  & \\ \hline
\cellcolor{Gray} MARSHA & \ding{51} & \ding{51} & \ding{51} & \ding{51} \\ \hline

\end{tabular}
\captionsetup{position=bottom}
\begin{flushleft}
\scriptsize \hspace{1.3cm} * Only those methods trained with humans in the shared workspace
\end{flushleft}
\caption{\rev{Capabilities of the proposed approach and other works in the literature.}}
\label{tab:quanlitative_comparison}

\end{table*}
\rev{
This Section summarizes the effectiveness of the methods covered in Section \ref{sec: related works}, including sampling-based path planners \cite{RRT*, Gammell:BIT, 9037111}, human-aware \cite{Casalino2019, Mainprice2011, Hayne2016} and safety-aware \cite{HAMP, S*, flowers2023spatiotemporal} proactive planners, reactive path replanners \cite{MARS, TPRM, DRRT, smart}, and learning-based approaches~\cite{Nicola2021, LUO2023104545, CHEN202264, 10576733}.
First, in Section~\ref{sec: target application}, we classify the methods according to their capabilities and describe a paradigmatic application to compare them. Then, in Section~\ref{sec: scenarios}, we assess the methods considering different interaction scenarios for the target application.
}

\subsection{Targeted Application} \label{sec: target application}

\rev{To evaluate the performance of the aforementioned methodologies, we consider the planner's capabilities:}
\begin{description}
    \item [Reactivity:] \rev{The ability to dynamically adjust the robot's path online based on the human's movements.}
    \item [Human-aware proactivity:] \rev{The ability to maintain human-robot distance or avoid areas typically occupied by humans.}
    \item [Safety-aware proactivity:] \rev{The ability to compute trajectories in line with ISO/TS 15066 safety guidelines.}
    \item [Robustness to changing working conditions:] \rev{The extent to which a planning pipeline adapts to new tasks or environments, indicating how challenging it is to apply the planner across different scenarios.}
\end{description}
\rev{Table \ref{tab:quanlitative_comparison} reports the classifications of the aforementioned methodologies according to such capabilities.
As a targeted application, we consider a cobot that palletizes boxes, and a human operator can access the workspace to inspect or move pallets.
The robot can move flexibly between its designated positions, while human interventions vary in duration and frequency, from brief, sporadic entries to a longer presence in the workspace. This setup often involves close human-robot proximity~\cite{KRUGER2009628}.
A safety perception system monitors the human’s position and communicates it to a safety module, which slows the robot down in compliance with ISO/TS 15066.}

\subsection{Scenarios}\label{sec: scenarios}

\subsubsection{Short and sporadic human presence} 
The human operator occasionally enters the robot's workspace unpredictably to briefly retrieve objects for inspection, triggering the robot's safety module and causing temporary slowdowns. Referring to Table \ref{tab:quanlitative_comparison}, algorithms that lack ``reactivity'' struggle in these situations. Proactive planning methods depend on predicting the operator's behavior through probabilistic models, which is difficult due to the infrequent entries.
In contrast, reactive planning algorithms can modify the robot's path in real-time when the operator is detected. 
As these human entries are short and infrequent, their impact on the robot's efficiency remains minimal, and simply relying on the safety module is often adequate for maintaining smooth operations.

\subsubsection{Short and frequent human presence} 
The human operator frequently enters the workspace briefly to feed the station with new boxes or take completed pallets away. These frequent entries often trigger the robot’s safety module, causing repeated slowdowns or complete stops, significantly reducing efficiency. 
As in the case of sporadic entries, proactive planning approaches fail to keep up with the unpredictability of frequent interventions.
Reactive planning algorithms can adjust the robot’s path but if they lack ``safety-awareness'' they may still lead to significant safety interventions. Frequent occurrences of these slowdowns can accumulate, negatively impacting productivity. An approach that combines reactive path adjustments with safety-aware strategies is more effective at minimizing disruptions and maintaining overall efficiency.

\subsubsection{Prolonged and structured human presence}

The human remains in the workspace for an extended time following predictable patterns, such as the robot placing boxes on a pallet while the operator closes each one. Deviations in the operator's movements are unlikely. This predictability enables the robot to plan its motions more effectively around the human's position. ``Human-aware'' proactive algorithms factor in the operator's position to compute paths that maintain a safe distance. For instance, learning-based approaches can learn policies that guide the robot to avoid proximity to humans, thereby minimizing the risk of collisions and reducing the need for frequent safety interventions. Incorporating ``safety-aware proactivity'' further optimizes the robot's travel time by reducing delays and ensuring smoother operations.
\subsubsection{Prolonged and unstructured human presence}
The operator stays in the workspace for an extended time with unpredictable movements. For instance, the operator may repeatedly move pallets or boxes between different locations within the workspace, with no fixed pattern, perhaps rearranging items to prepare for inspections or transfers. Here, even ``safety-aware'' algorithms lacking ``reactivity'' capabilities fail to prevent the robot from long-term potential safety stops when unexpected obstructions occur.  These algorithms cannot cope with human movement which significantly deviates from expected patterns. 

\subsubsection{Changes to task or environment}
\rev{The robot's tasks, workspace layout, or parameters may change frequently in highly dynamic environments, such as those typical in high-mix, low-volume manufacturing. Examples include relocating pick-up areas, accommodating obstacles like accumulated parts, integrating new operators, or adapting to different tasks as production fluctuates.
Traditional sampling-based planners adjust well to such conditions, leveraging vision system inputs to respond to new configurations. However, learning-based methods, often pre-trained for specific robots and fixed environments, may require substantial adaptation retraining, which is impractical for rapidly changing setups.} 

\begin{figure}[t]
    \centering
    \includegraphics[width=0.90\columnwidth]{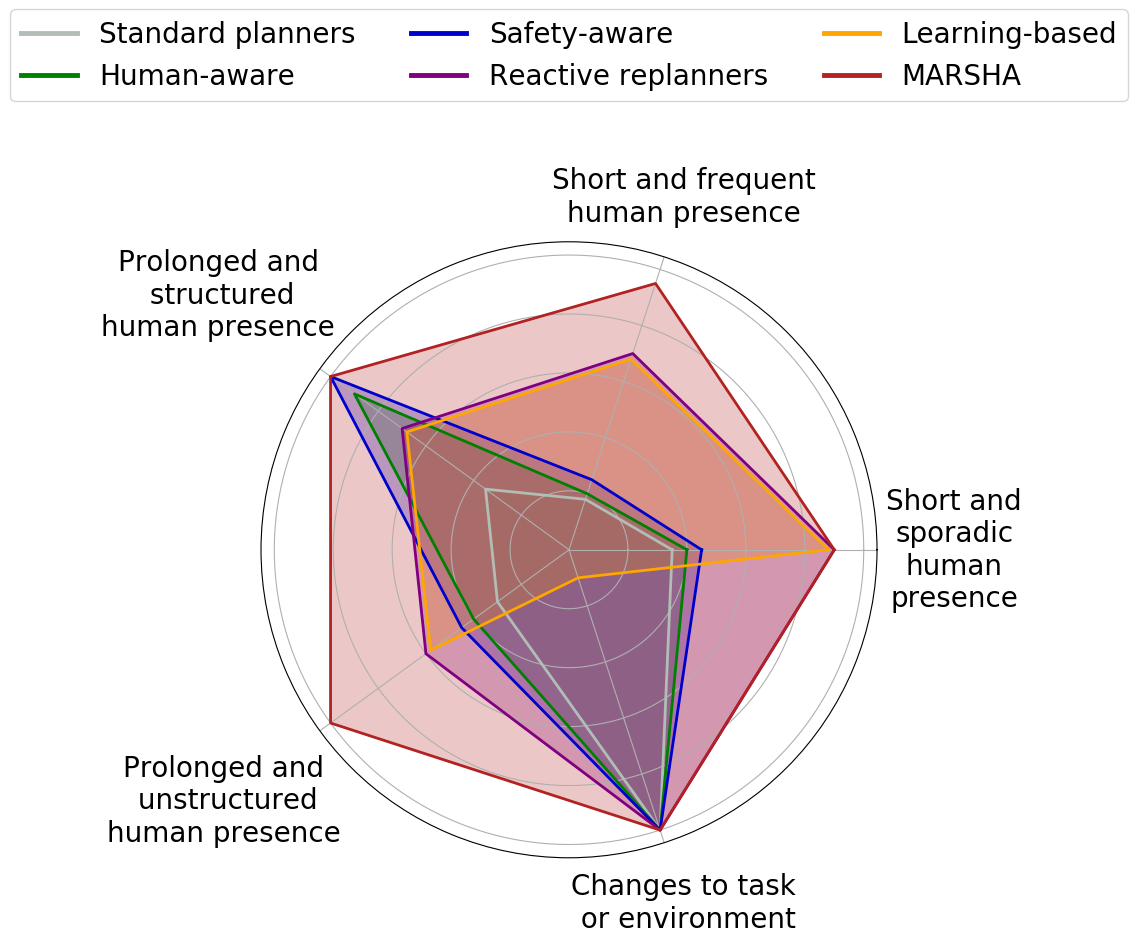} 
    \caption{\rev{Comparison of algorithm categories listed in Table \ref{tab:quanlitative_comparison} considering the scenarios in Section \ref{sec: qualitative_assessment}. Each scenario assumes a safety module based on ISO/TS 15066’s SSM guidelines. The closer an algorithm's shape extends to the outer perimeter of the chart, the better it performs.}}
    \label{fig:radar_chart}
\end{figure}

\rev{The performance of the various algorithms in the described scenarios is illustrated in Fig.  \ref{fig:radar_chart}. MARSHA promises strong performance across all scenarios due to its combination of proactive, safety-aware planning and reactive capabilities and inherent adaptability to dynamic tasks and changing environments.}

\section{Experiments} \label{sec: case_study}
\rev{We experimentally compare MARSHA with reactive and proactive approaches, prioritizing those that share MARSHA’s ability to adapt to varying working conditions.}

\subsection{Selection of Reactive Approaches Rationale}
We evaluated two approaches in comparison to MARSHA. The first approach, dSSM (dynamic SSM 2D \cite{BYNER2019239}), continually monitors both human and robot statuses, determining the maximum allowable robot speed based on  \eqref{eq: vmax SSM}. If the robot's speed exceeds the maximum allowed by safety considerations, it is scaled by a factor $\lambda$. 
Concerning Fig. \ref{fig: pipeline}, there is no replanning block. The second reactive approach used in the experiments combines dSSM with a path replanning algorithm (MARS \cite{MARS}). In the first case, the robot slows down according to the safety criterion defined by  \eqref{eq: vmax SSM}. In the second case, the robot has the additional capability to replan its path. 
The dSSM approach holds the robot as long as the operator is nearby, while MARS would modify the path to find collision-free solutions but at significantly reduced speed. It is worth noting that MARS modifies the path only when encountering an obstacle or identifying a shorter solution. 
Instead, MARSHA predicts the effect of the safety module on the solutions found and aims to minimize the traversing time, considering the speed slowdowns due to human movements. MARS and MARSHA were assigned a maximum calculation time of $200$ ms\footnote{As per Algorithm \ref{alg: threads}, the robot tracks the trajectory at a high rate (\textit{e.g.,} $500$ Hz). In contrast, trajectory updates occur at a rate up to $5$ Hz.}.

\subsection{Selection of Proactive Approaches Rationale}
For proactive methods, MARSHA was compared to a safety-aware path planning algorithm (HAMP \cite{HAMP}) and a minimum-length path planning algorithm (MIN-LEN), where the human's expected volume is treated as an obstacle.
\rev{The MIN-LEN algorithm refers to an optimal path planning algorithm with sufficient computational time to practically achieve the optimal solution, such as RRT* \cite{RRT*}, BIT* \cite{Gammell:BIT}, or other optimal planners (e.g., \cite{9037111,MIRRT})}\footnote{\rev{These experiments focus only on the time required to \textit{execute} a trajectory in a collaborative scenario, not in the offline computation time.}}.
%
Both approaches were integrated with dSSM during runtime.
HAMP aims to find a time-optimal solution by minimizing \eqref{eq: hamp_cost_fcn3}, generating paths that require minimal or no intervention from the safety module as long as the operator remains within the expected space defined during the planning phase. However, HAMP's performance deteriorates if the operator deviates from this space.
Conversely, MIN-LEN seeks to identify the shortest path while treating the human as an obstacle. This approach  results in shorter paths but at the expense of more frequent safety module activations.
Referring to Fig. \ref{fig: pipeline}, tests with HAMP and MIN-LEN do not implement replanning blocks.
%


\subsection{Design of Experiments and Metrics} \label{sec: desing_exp}
We evaluate the following metrics: the \textit{Trajectory execution time}, which measures the time taken to complete the task, normalized relative to the minimum execution time (\textit{i.e.,} the time obtained when executing the trajectory without human presence), and the \textit{Average scaling factor} (ranging from 0 to 100), which represents the intervention of the safety module. A lower average scaling factor indicates a higher level of intervention by the safety module in reducing the robot's speed.

\begin{figure}[t]
    \centering
    \subfloat[Real robotic cell.\label{fig: real_env}]{%
        \includegraphics[height=3cm]{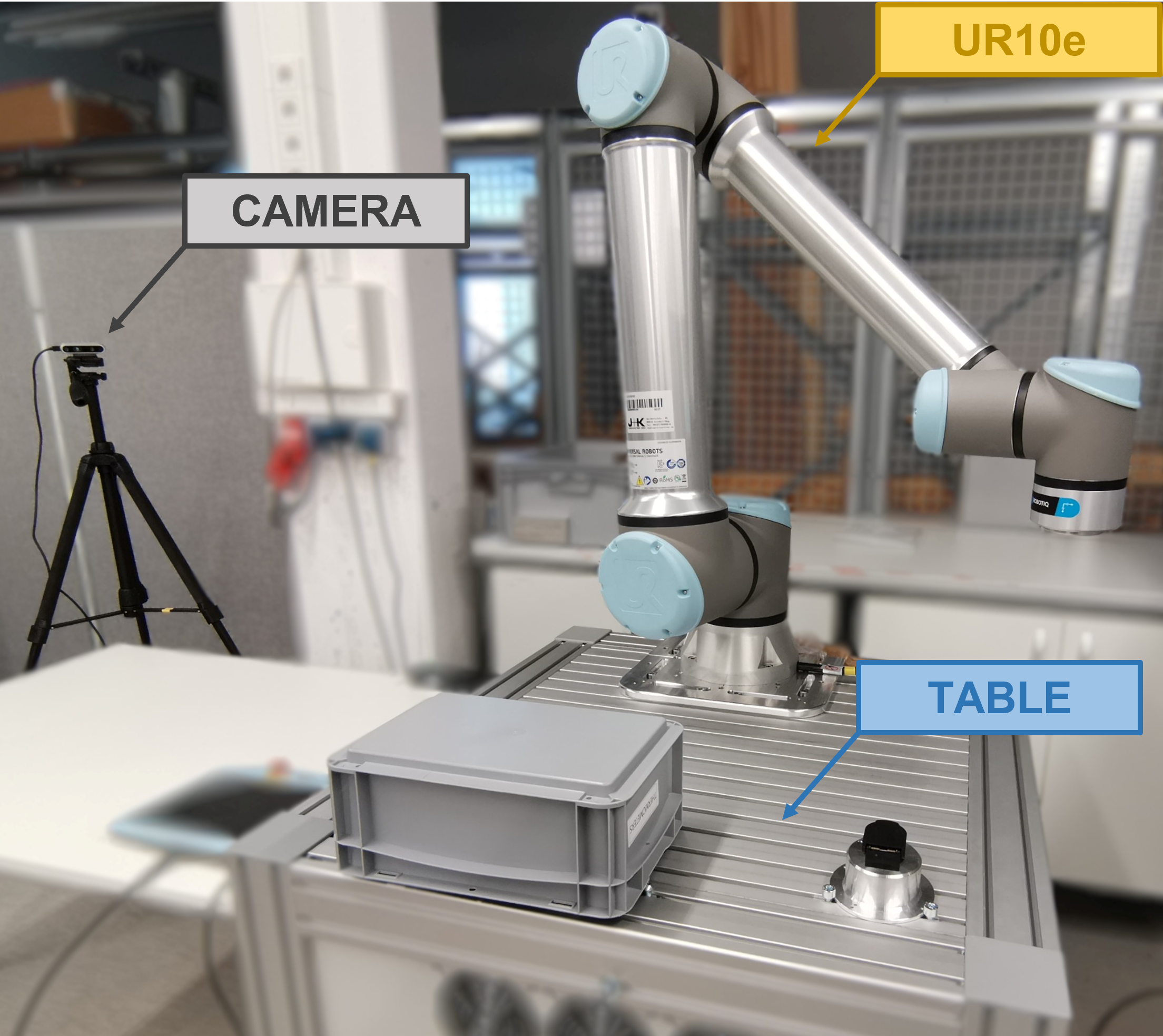}
    }
    \hfill
    \subfloat[Simulated environment. \label{fig: simulation_env}]{%
      \includegraphics[height=3cm,trim={0 1cm 0 1cm},clip]{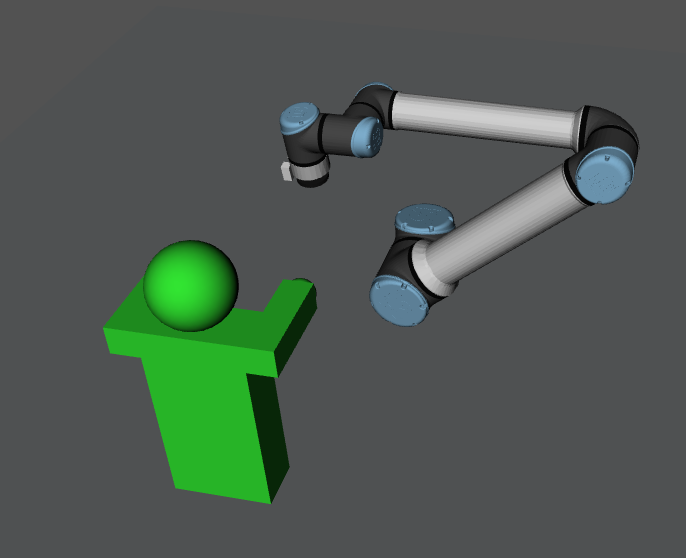}
    }
    \caption{Experimental setups.}
\label{fig: experimental_ setup}
\end{figure}

We conduct simulated and real-world experiments.
The setup  (Fig. \ref{fig: real_env}) comprises a Universal Robot UR10e,  a shared table, and an Intel RealSense D435. We compute the human skeleton bounding boxes and feed them into the ISO/TS 15066 computation. 
In simulations, we replicate the human movements using a mannequin (Fig. \ref{fig: simulation_env}) with added uniform noise (max. amplitude of $3$ cm). The locations of the dummy's head, torso, arms, and hands are used for the safety computation. 
During real experiments, we limit the robot's speed to $30\%$ of its maximum for safety reasons. SSM implements \eqref{eq: vmax SSM}  with $T_r = 0.15 ~s$, $a_s = 2.5 ~m/s^2$, $C = 0.25 ~m$ and $v_h = 1.6 ~m/s$.

The control pipeline follows Fig. \ref{fig: pipeline}.
MARSHA and the reactive methods are implemented in the ``fast safety-aware replanner'' module, while the proactive methods are in the ``offline path planner'' module.
The code is in C++ and executed in  ROS \cite{ROS}. The replanning algorithm were implemented in OpenMORE \cite{openmore}, an open-source library for robot motion replanning. 

The tasks consist of a workspace-sharing application. The robot's goal is to move from an initial configuration to a designated target configuration to execute tasks like pick\&place operations. 
Concurrently, an operator approaches and works on the same table. 
%


\subsection{Comparison with Reactive Approaches}
\subsubsection{Simulation} \label{sec: reactive_test}
\begin{figure*}
    \centering
    \subfloat[ ]{%
        \includegraphics[width = 0.235\columnwidth]{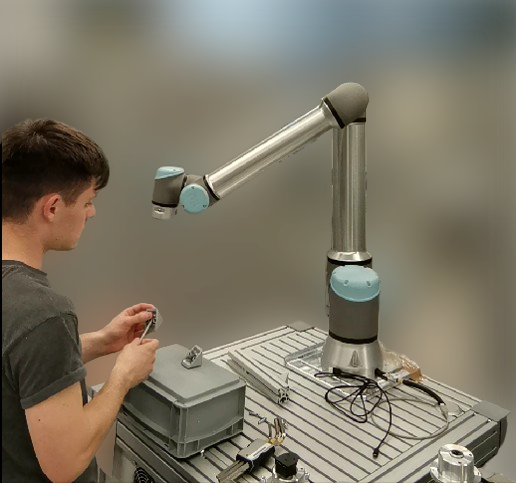}
    }
    \hfill
    \subfloat[ ]{%
        \includegraphics[width = 0.235\columnwidth]{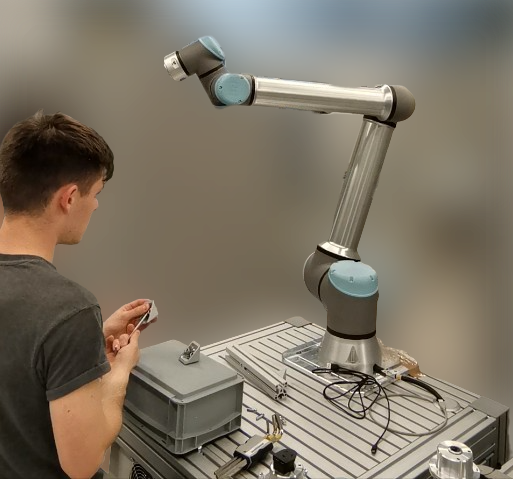}
    }
     \hfill
    \subfloat[ ]{%
        \includegraphics[width = 0.235\columnwidth]{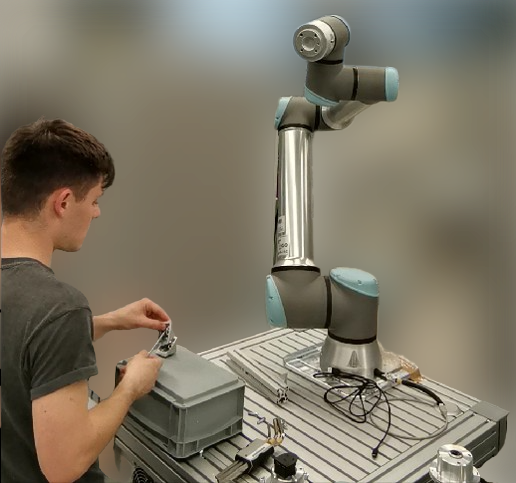}
    }
     \hfill
    \subfloat[ ]{%
        \includegraphics[width = 0.235\columnwidth]{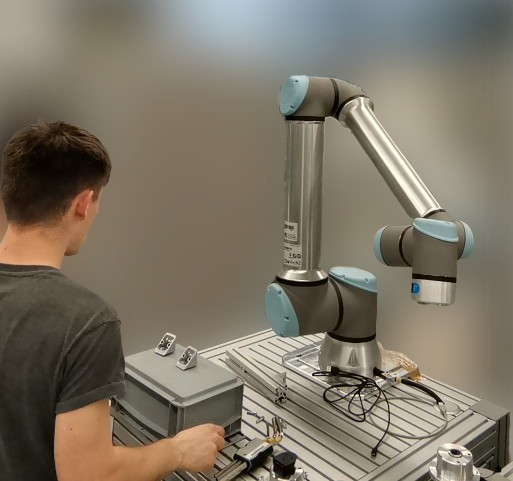}
    }
    \hfill
    \subfloat[ ]{%
        \includegraphics[width = 0.235\columnwidth]{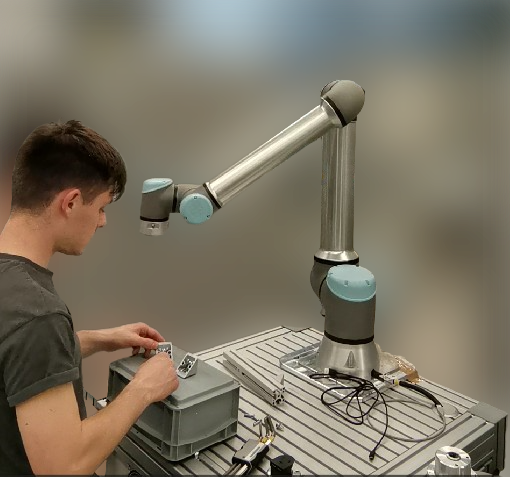}
    }
    \hfill
    \subfloat[ ]{%
        \includegraphics[width = 0.235\columnwidth]{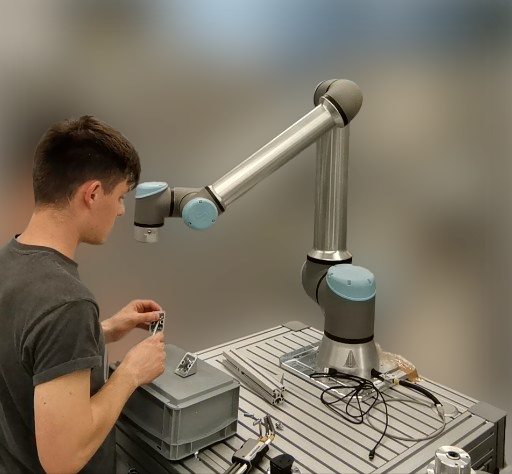}
    }
     \hfill
    \subfloat[ ]{%
        \includegraphics[width = 0.235\columnwidth]{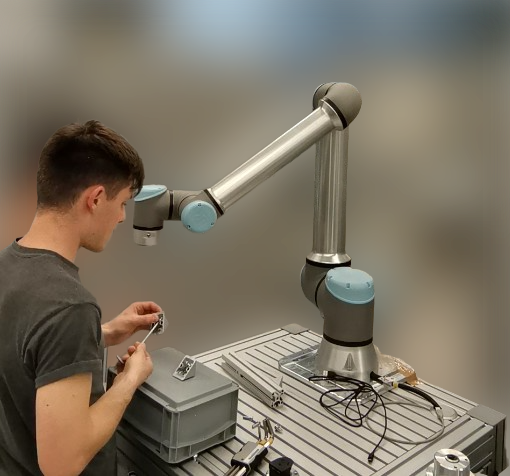}
    }
     \hfill
    \subfloat[]{%
        \includegraphics[width = 0.235\columnwidth]{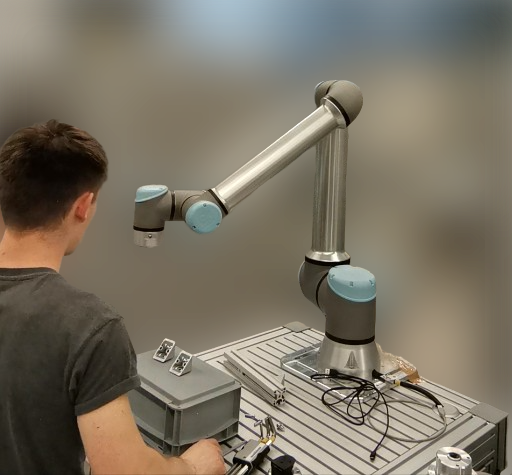}
    }
    \caption{(a)-(d) MARSHA enables the robot to dynamically adjust its path in response to the operator's approach. 
    (e)-(h) dSSM halts the robot as long as the operator is nearby.}
\label{fig: screenshots from tests}
\end{figure*}
The experiments primarily varied based on the duration of the operator's presence at the workbench. 
The tests were carried out as follows. The robot plans a path from start to goal while the operator is still far away. The trajectory initially corresponds to the fastest one in the absence of the operator. Two additional paths are computed for MARS and MARSHA, as discussed in Section \ref{sec: replanning}. As the robot moves, the operator approaches it to work on the shared table. Depending on the algorithm used, the robot slows down or replans its path. After a pre-established time interval, the operator leaves.

We consider three cases: \textit{short}, \textit{medium}, and \textit{long}, which are distinguished by the time during which the operator is close to the robot, which is $5$, $10$ and $20$ seconds, respectively. $50$ simulations and $20$ experiments on the real cell were performed for each type of test and algorithm.
Fig \ref{fig: screenshots from tests} shows snapshots of MARSHA and dSSM running during the experiments.
\begin{figure*}
\centering
    \captionsetup[subfloat]{labelformat=empty}
    \subfloat[]{%
        \includegraphics[height = 0.18\textwidth]{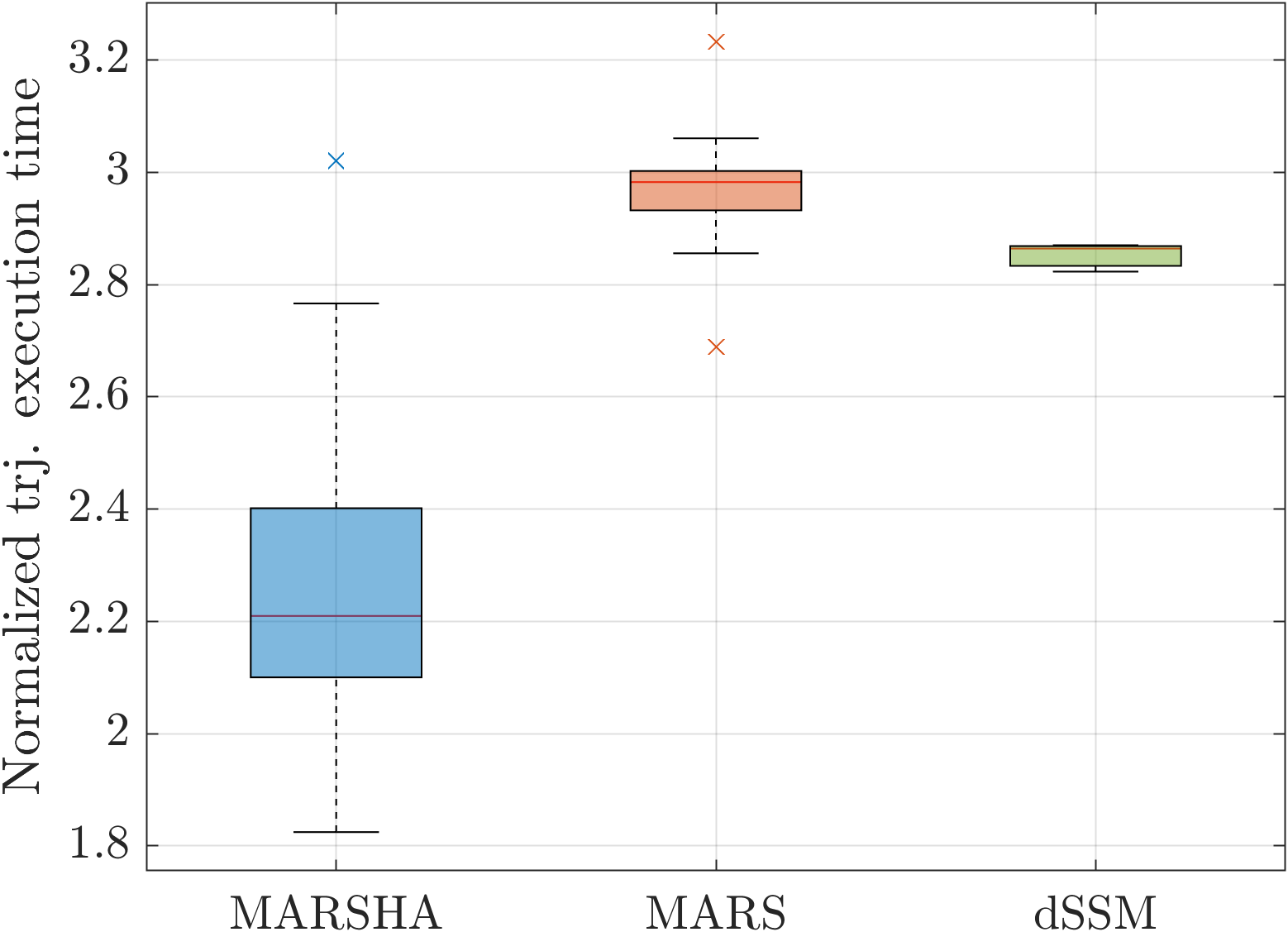}
    }\qquad
    \subfloat[]{%
        \includegraphics[height = 0.18\textwidth]{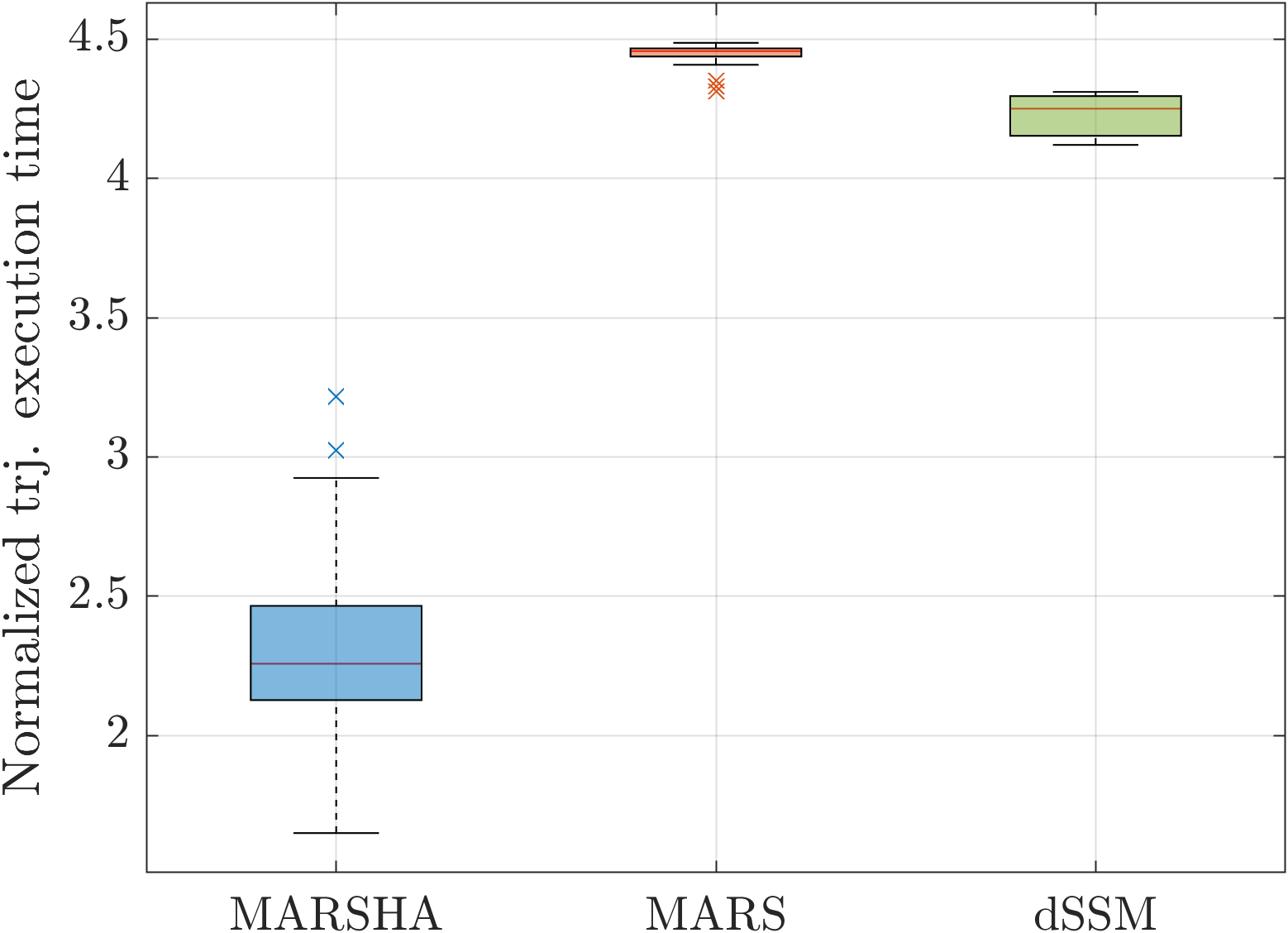}
    }\qquad\,
    \subfloat[ ]{%
        \includegraphics[height = 0.18\textwidth]{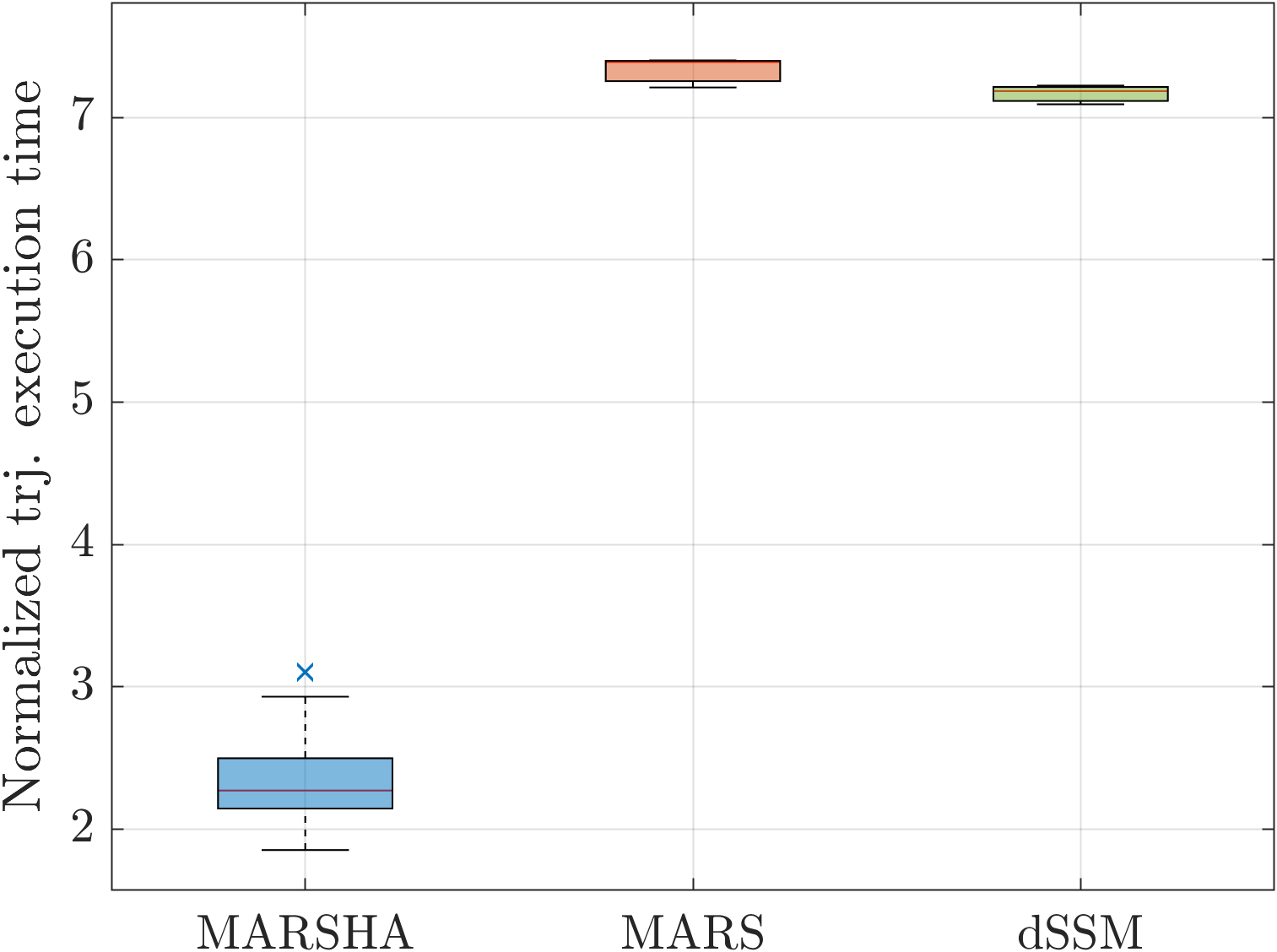}
    }
    \\
    \vspace{-0.5cm}
    \!\!
    \subfloat[(a)]{%
        \includegraphics[height = 0.18\textwidth]{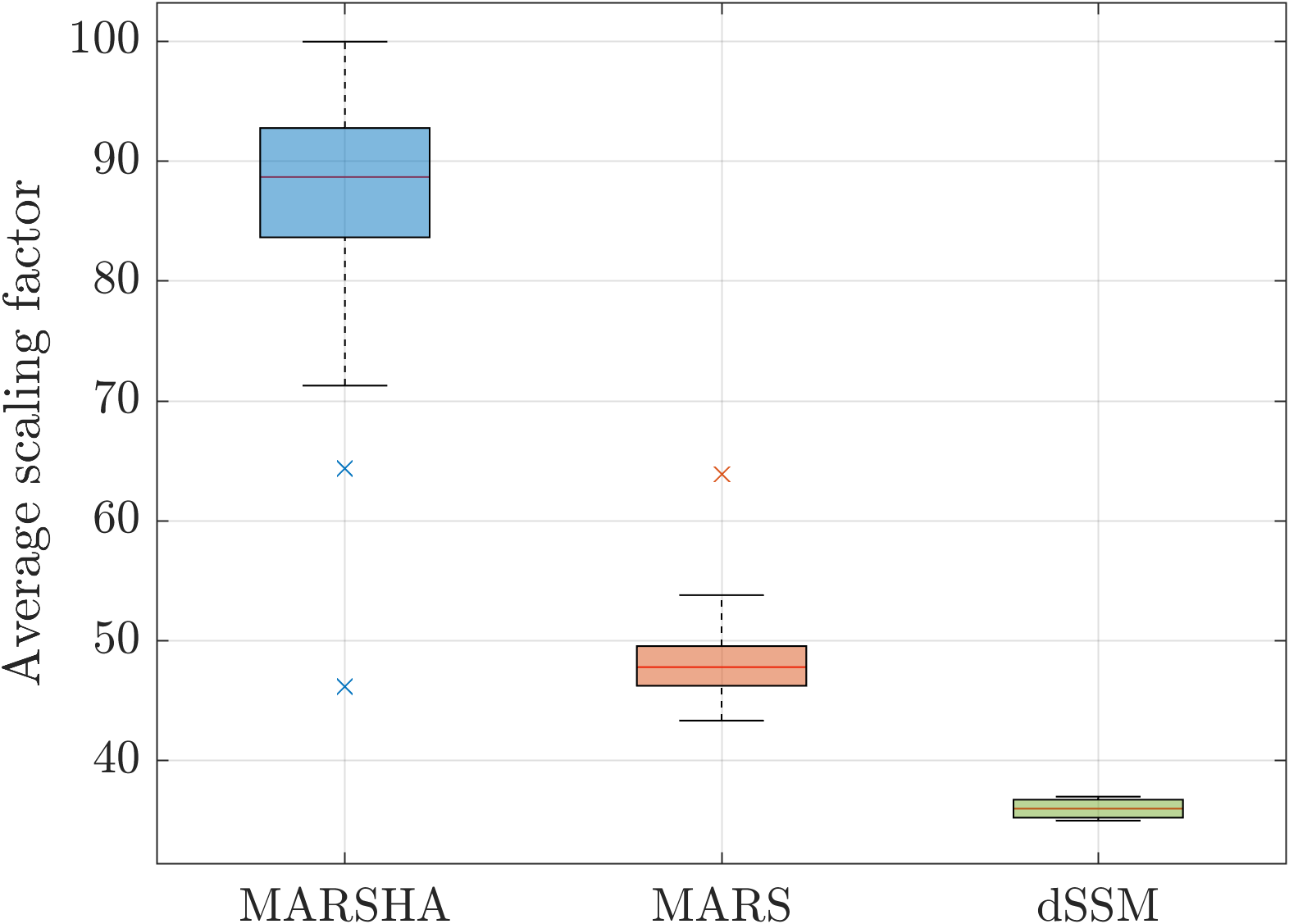}
    }\qquad
    \subfloat[(b)]{%
        \includegraphics[height = 0.18\textwidth]{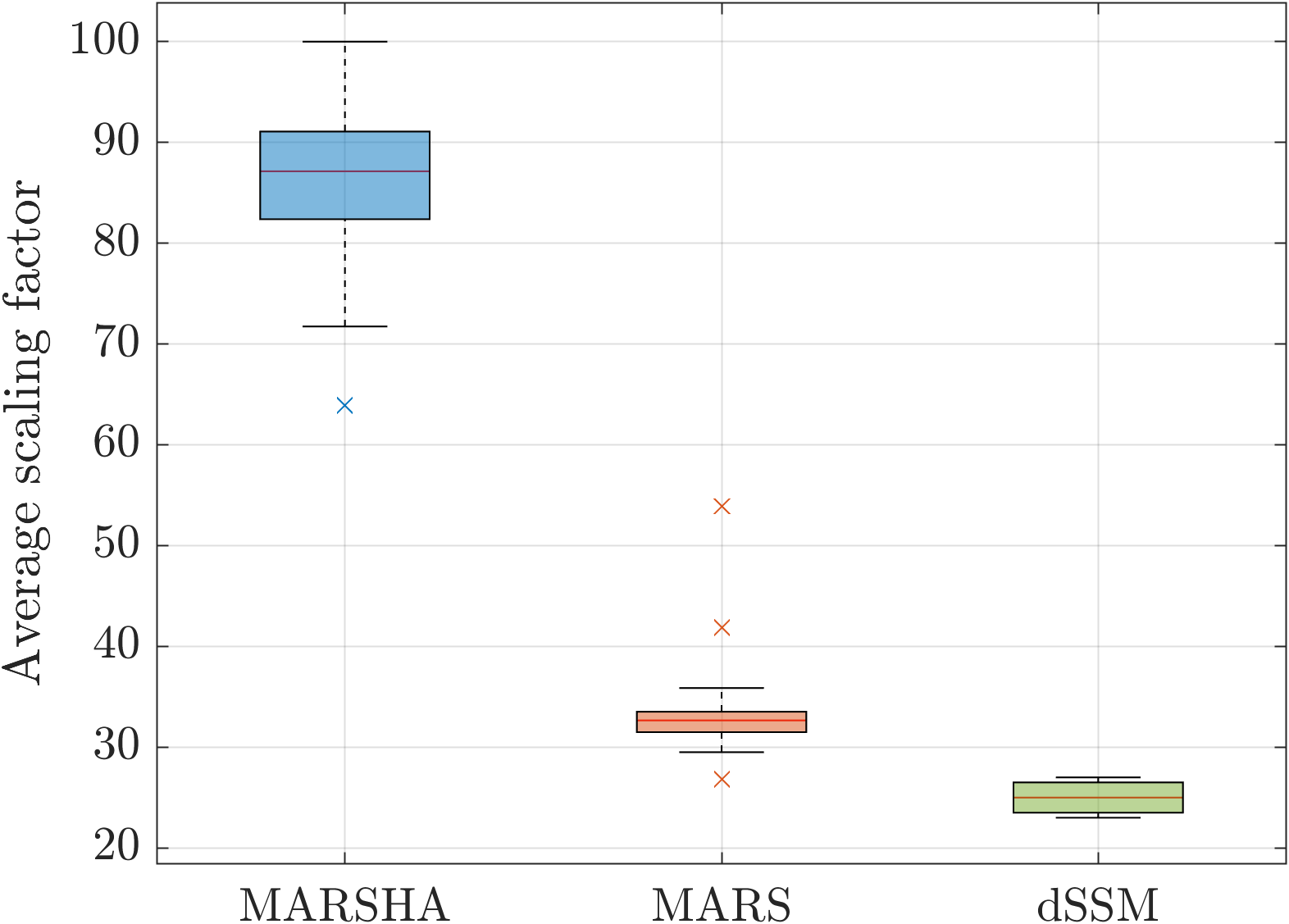}
    }\qquad
    \subfloat[(c)]{%
        \includegraphics[height = 0.18\textwidth]{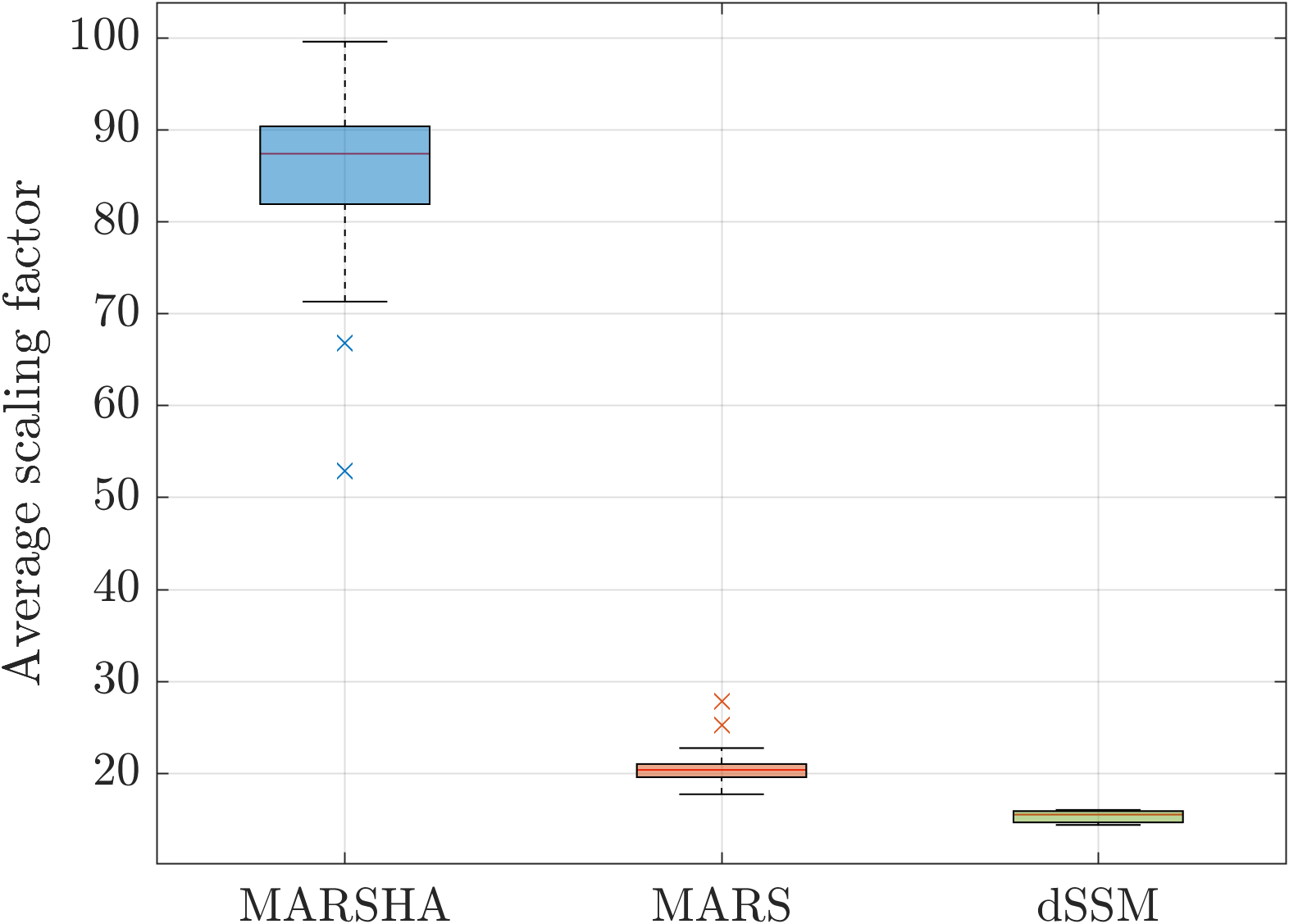}
    }
\caption{Simulation comparison with reactive approaches. (a) \textit{short} interaction ($5$ seconds);
(b) \textit{medium} interaction ($10$ seconds);
(c) \textit{long} interaction ($20$ seconds).}
\label{fig: simulated_reactive_test}
\end{figure*}
Fig. \ref{fig: simulated_reactive_test} depicts the outcomes of the \textit{short}, \textit{medium}, and \textit{long} tests conducted in the simulation.
A lower trajectory execution time and a higher average scaling factor indicate superior algorithm performance. \rev{Upon analysis, it becomes evident that MARSHA outperforms both MARS and dSSM in all tested scenarios, reducing the trajectory execution time by approximately $60\%$ in \textit{long} tests}. MARSHA can generate path plans that reduce activations of the safety module, enabling the robot to reach its goal more expeditiously. This behavior remains consistent across all three test variations, regardless of the operator's duration of interaction with the workbench.
Conversely, MARS and dSSM exhibit poor performance when the operator remains near the robot for an extended period. dSSM restricts the robot's movement for the entire duration of the operator's presence nearby. MARS's replanning feature enhances safety by ensuring collision-free paths but compromises the achieved performance. Since MARS continuously finds and optimizes new paths in terms of length, the solutions frequently bring the robot near the operator, eliciting strong responses from the safety module. Both MARS and dSSM result in low average scaling factors, indicating that the robot remains mostly stationary during the tests. In contrast, MARSHA exhibits a higher average scaling factor, reflecting fewer safety module interventions. The higher variability of MARSHA scaling factor arises from the stochastic nature of the sampling-based approach and test conditions (\textit{e.g.,} different timing of the human movements).
The performance gap between MARSHA and MARS/dSSM is expected to widen as the operator's time at the robot increases. Code to run a  simulation is available at \cite{marsha_examples}.

\subsubsection{Real scenario} \label{sec: reactive_test_real}
\begin{figure*}
\centering
    \captionsetup[subfloat]{labelformat=empty}
    \subfloat[]{%
        \includegraphics[height = 0.18\textwidth]{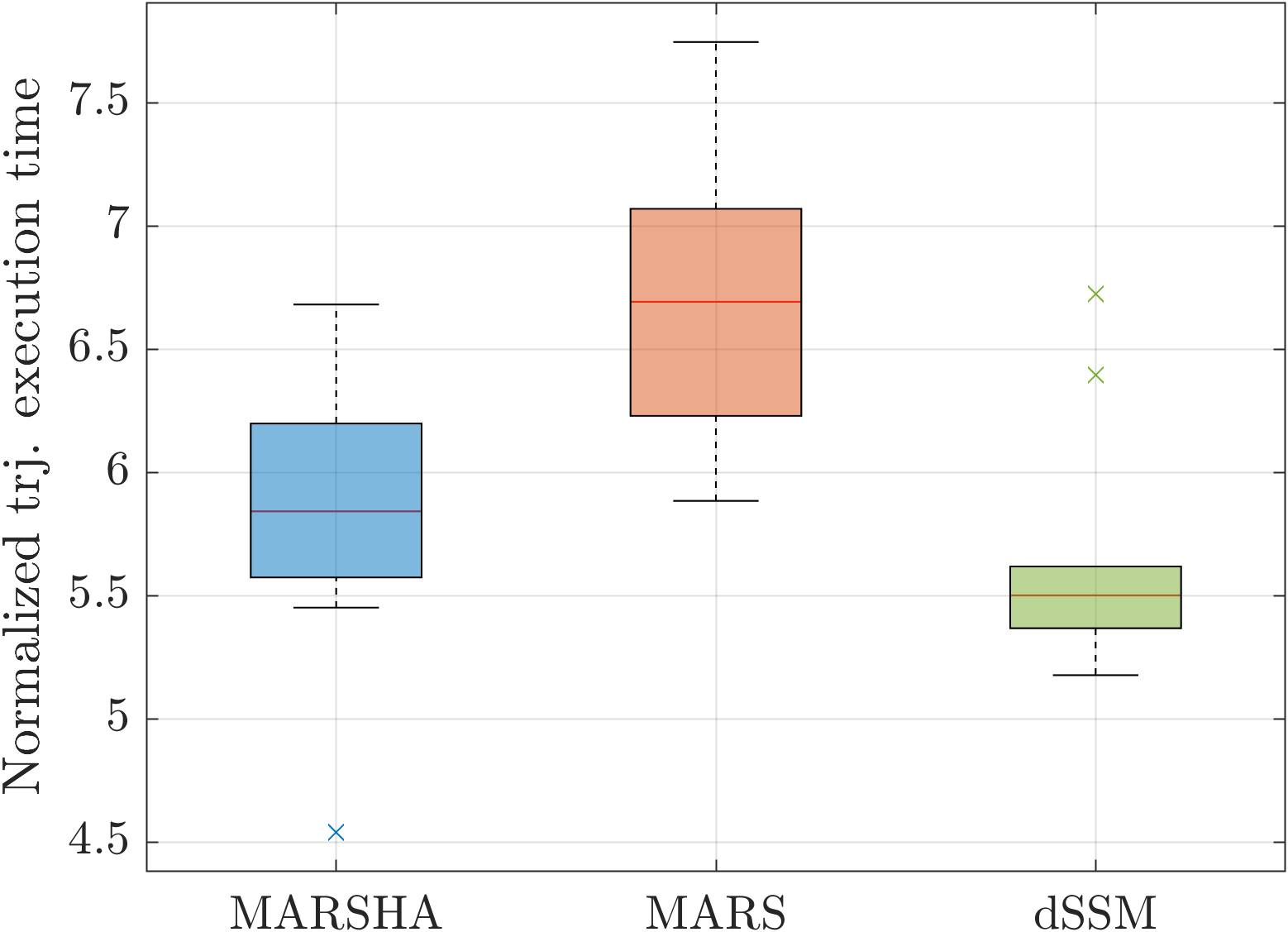}
    }\qquad
    \subfloat[]{%
        \includegraphics[height = 0.18\textwidth]{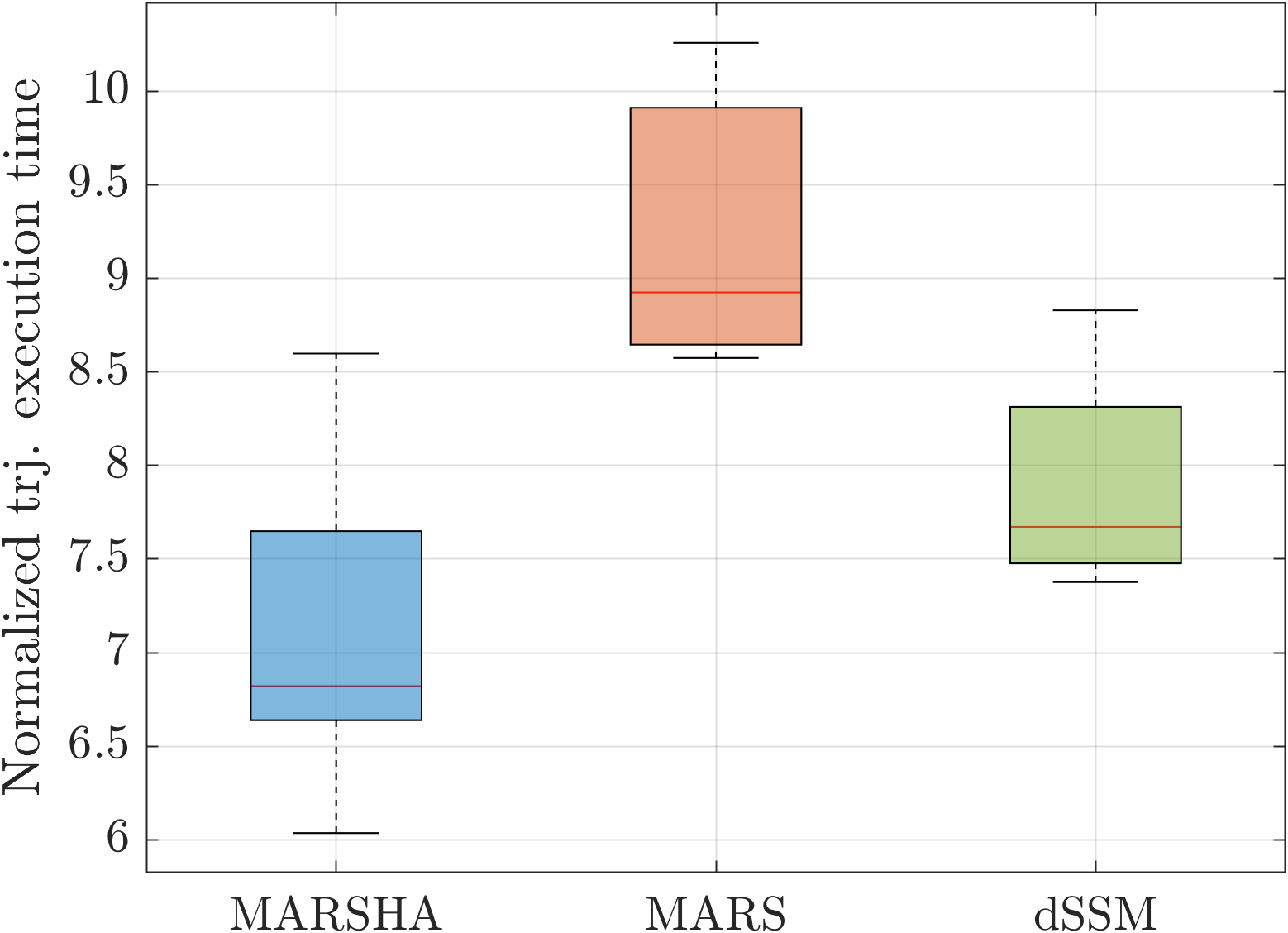}
    }\qquad\,
    \subfloat[ ]{%
        \includegraphics[height = 0.18\textwidth]{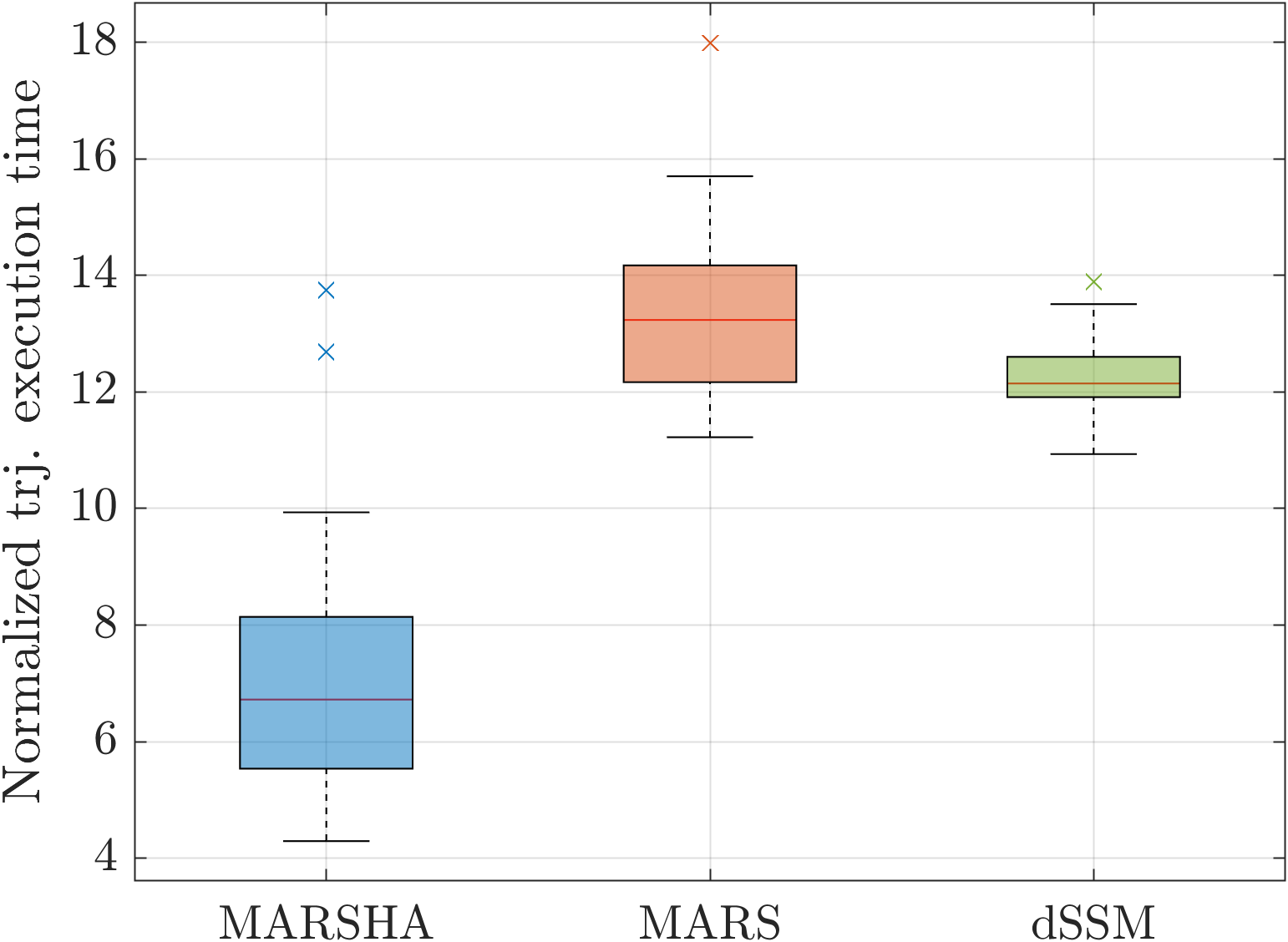}
    }
    \\
    \vspace{-0.5cm}
    \!\!
    \subfloat[(a)]{%
        \includegraphics[height = 0.18\textwidth]{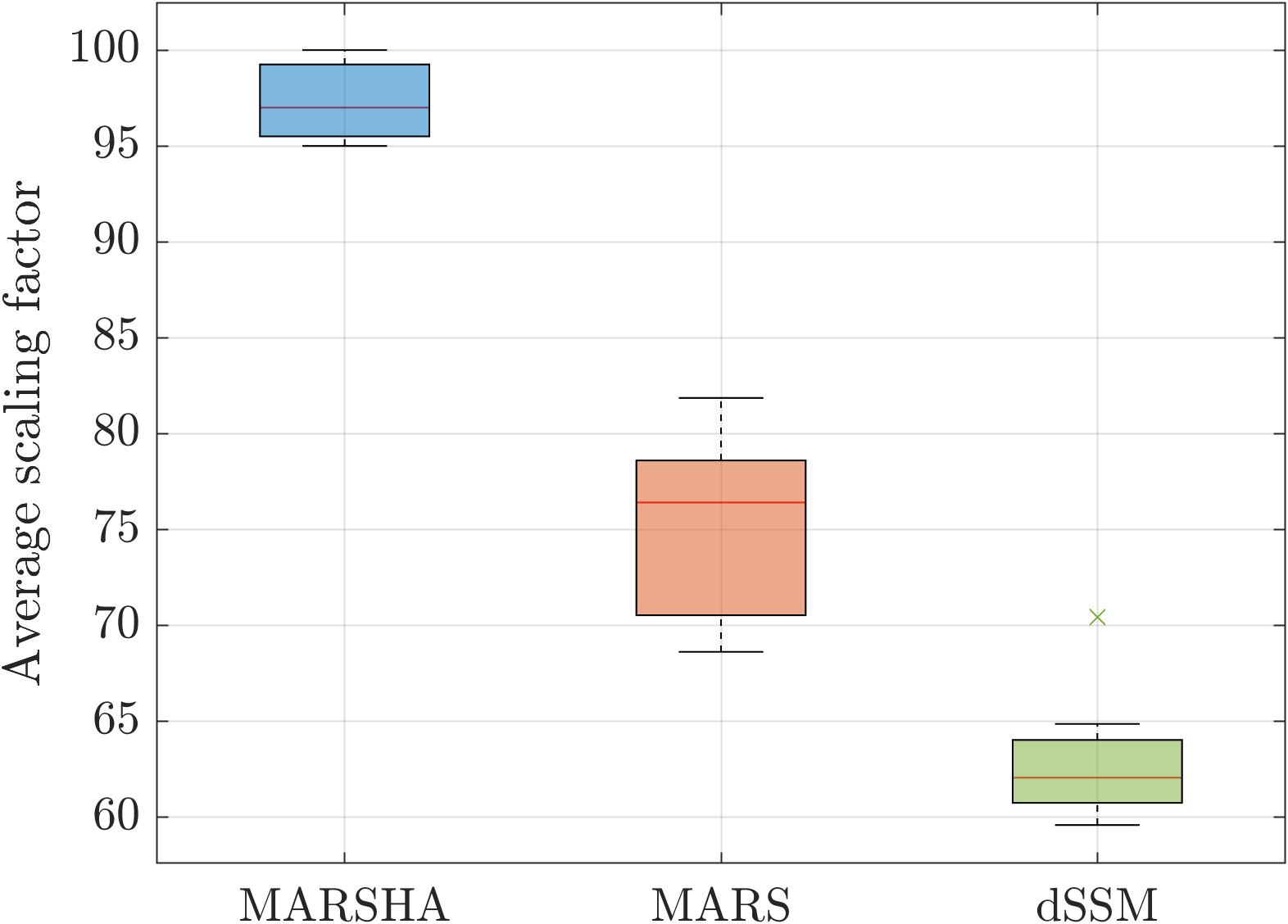}
    }\qquad
    \subfloat[(b)]{%
        \includegraphics[height = 0.18\textwidth]{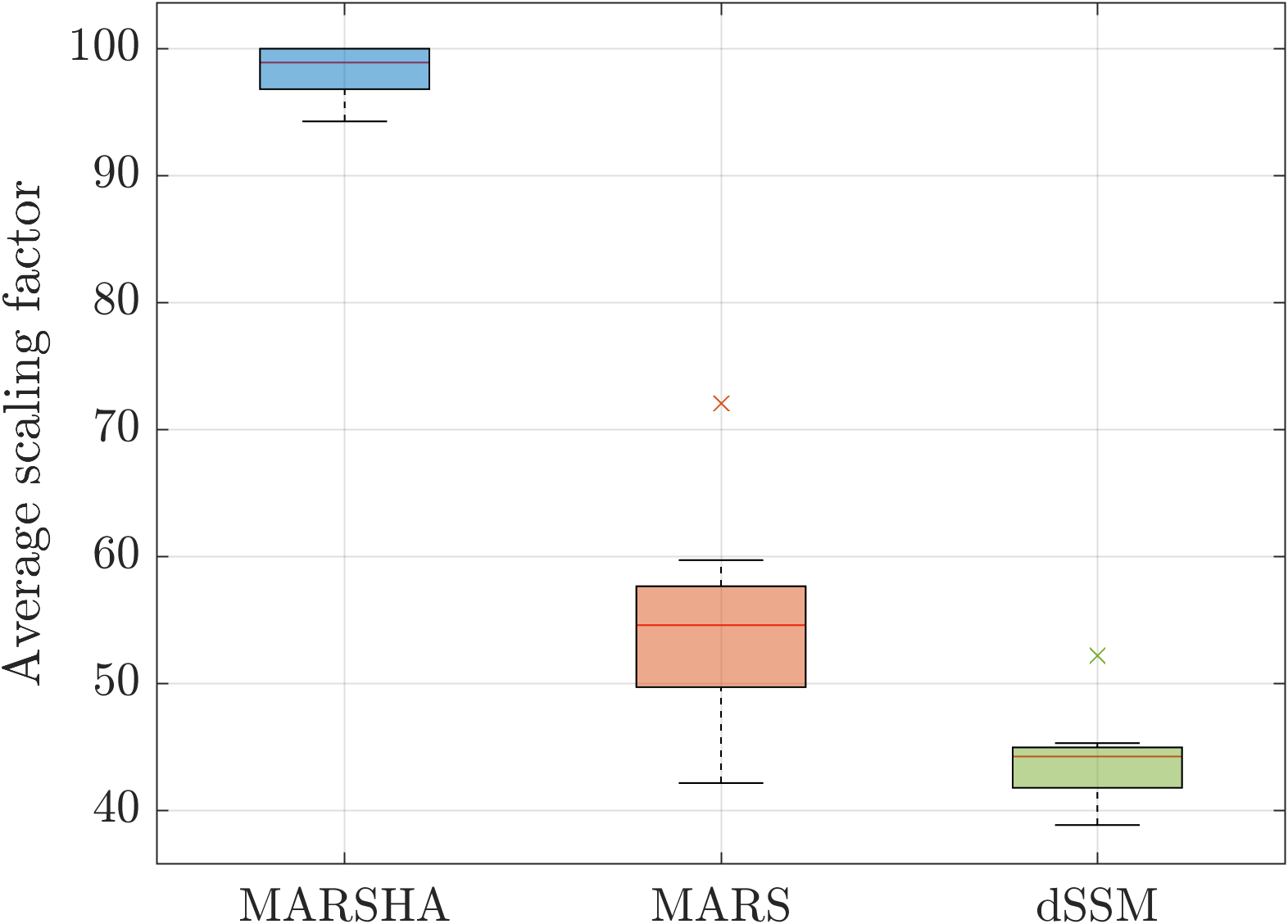}
    }\qquad
    \subfloat[(c)]{%
        \includegraphics[height = 0.18\textwidth]{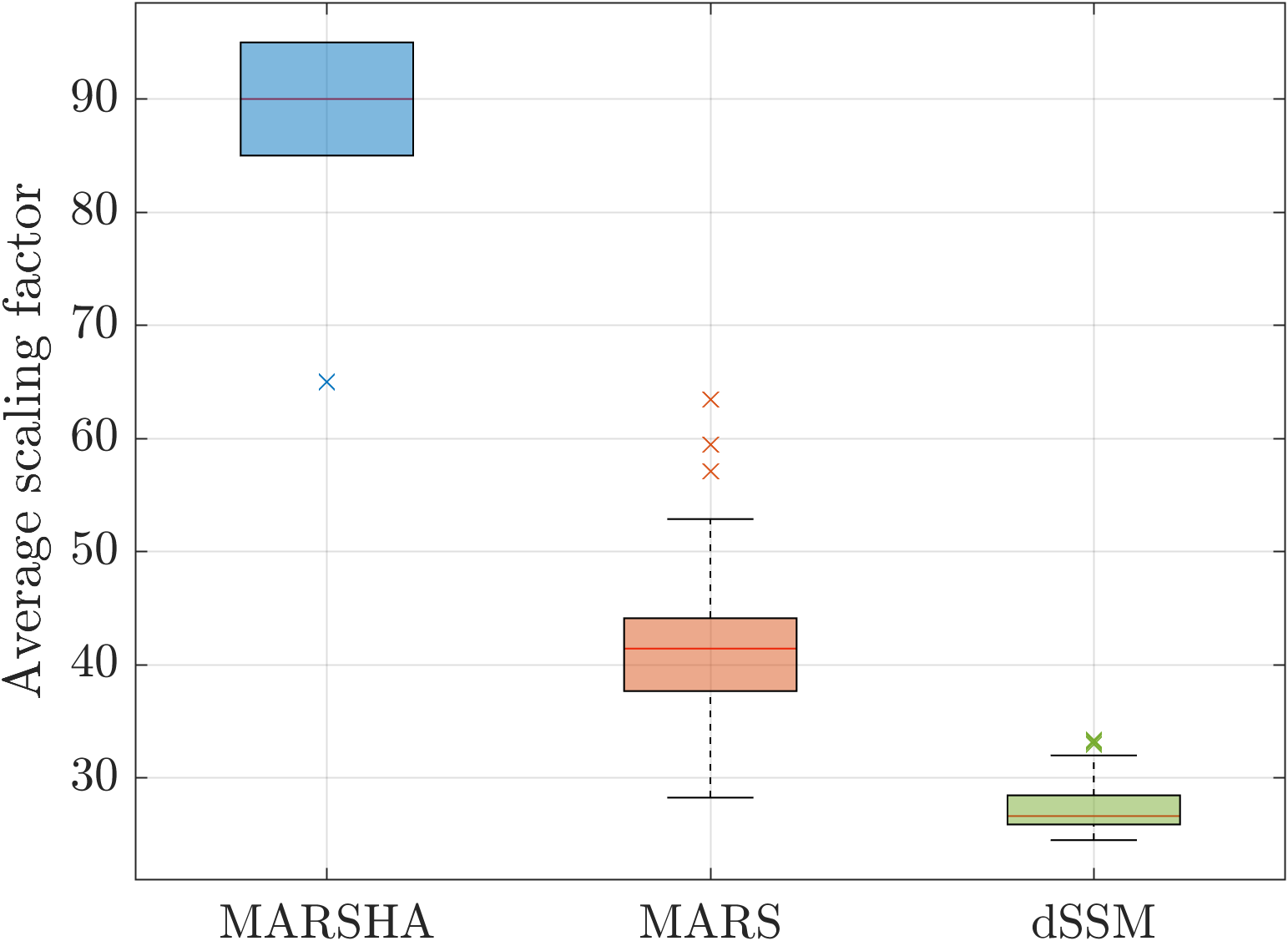}
    }
\caption{Real experiments with reactive approaches. (a) \textit{short} interaction ($5$ seconds);
(b) \textit{medium} interaction ($10$ seconds);
(c) \textit{long} interaction ($20$ seconds).}
\label{fig: real_reactive_test}
\end{figure*}
The findings from the simulations are further validated through experiments conducted on the physical robot, as in Fig. \ref{fig: real_reactive_test}. 
The results show that MARSHA can quickly identify executable paths by considering the impact of the safety module on the robot during execution. As the duration of the operator's presence in the shared space increases, MARSHA performs better than MARS/dSSM. 
However, the performance disparity between the algorithms narrows. This can be attributed to the robot operating at $30\%$ of its total capacity during real-world tests. Consequently, the distance the robot covers before the operator departs is reduced. As a result, the robot often does not reach the portion of the path replanned by MARSHA before the operator leaves the workspace. In this way, MARSHA's advantages are not fully utilized, leading to a comparable performance with dSSM as the robot's maximum speed decreases and the duration of the operator's presence at the workstation diminishes.
To confirm this hypothesis, we conducted simulated \textit{short} tests with constant scaling at $30\%$ and $60\%$, combined with the scaling provided by the safety module. As depicted in Fig. \ref{fig: speed_scaled}, reducing the robot's speed corresponds to a smaller performance gap between the algorithms. When the robot's speed is restricted, and the operator's interventions are brief and infrequent, MARSHA performs as dSSM. Additionally, other factors, such as the positioning of the operator and the timing of entry into the cell, may influence the results, as they are not always the same during real-world tests.
\begin{figure}
        \centering
        \captionsetup[subfloat]{labelformat=empty}
        {%
        \includegraphics[height = 3.6 cm]{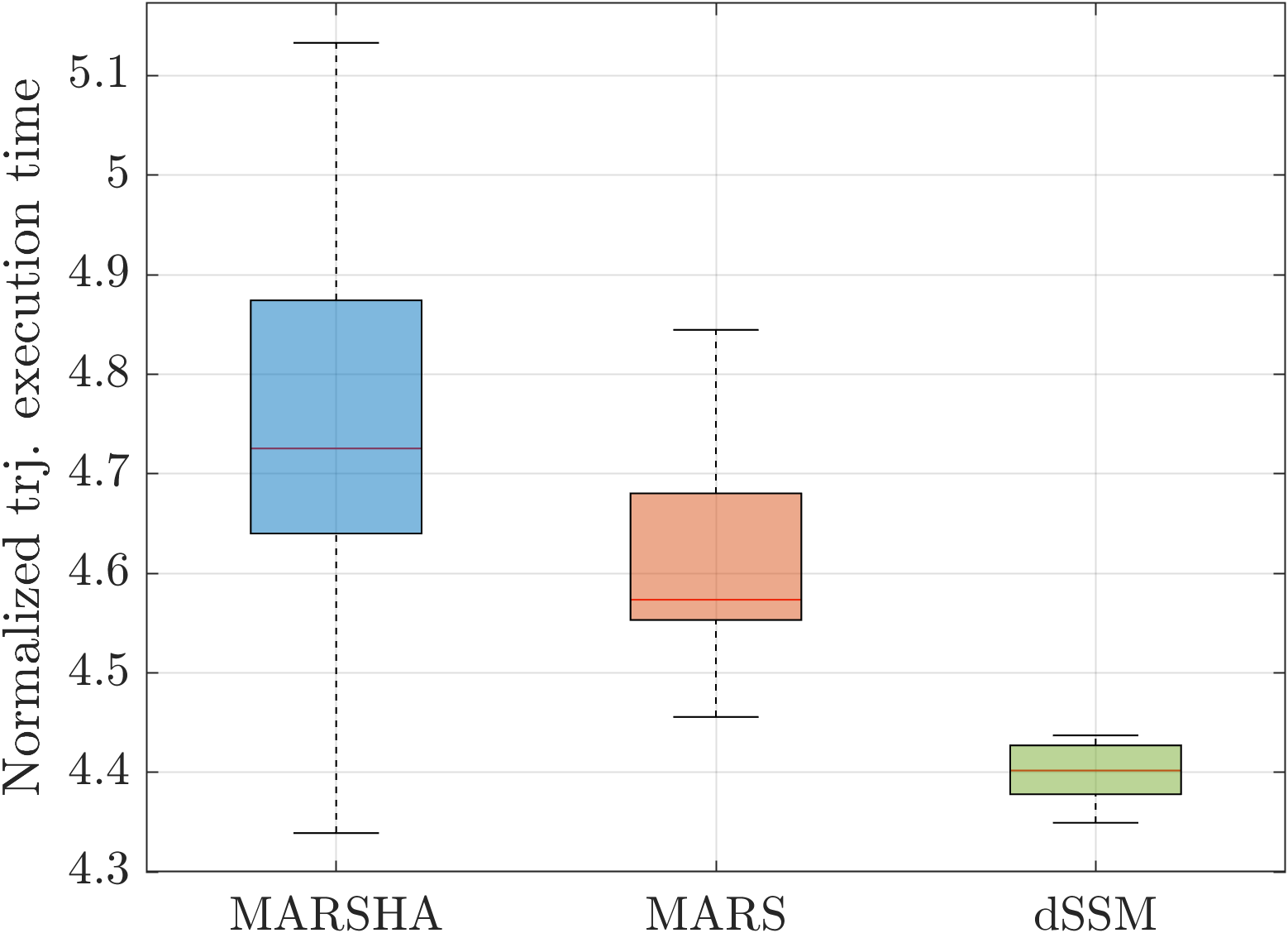}
    }
    \\
    \vspace{0.2cm}
    \!\!
        {%
        \includegraphics[height = 3.6 cm]{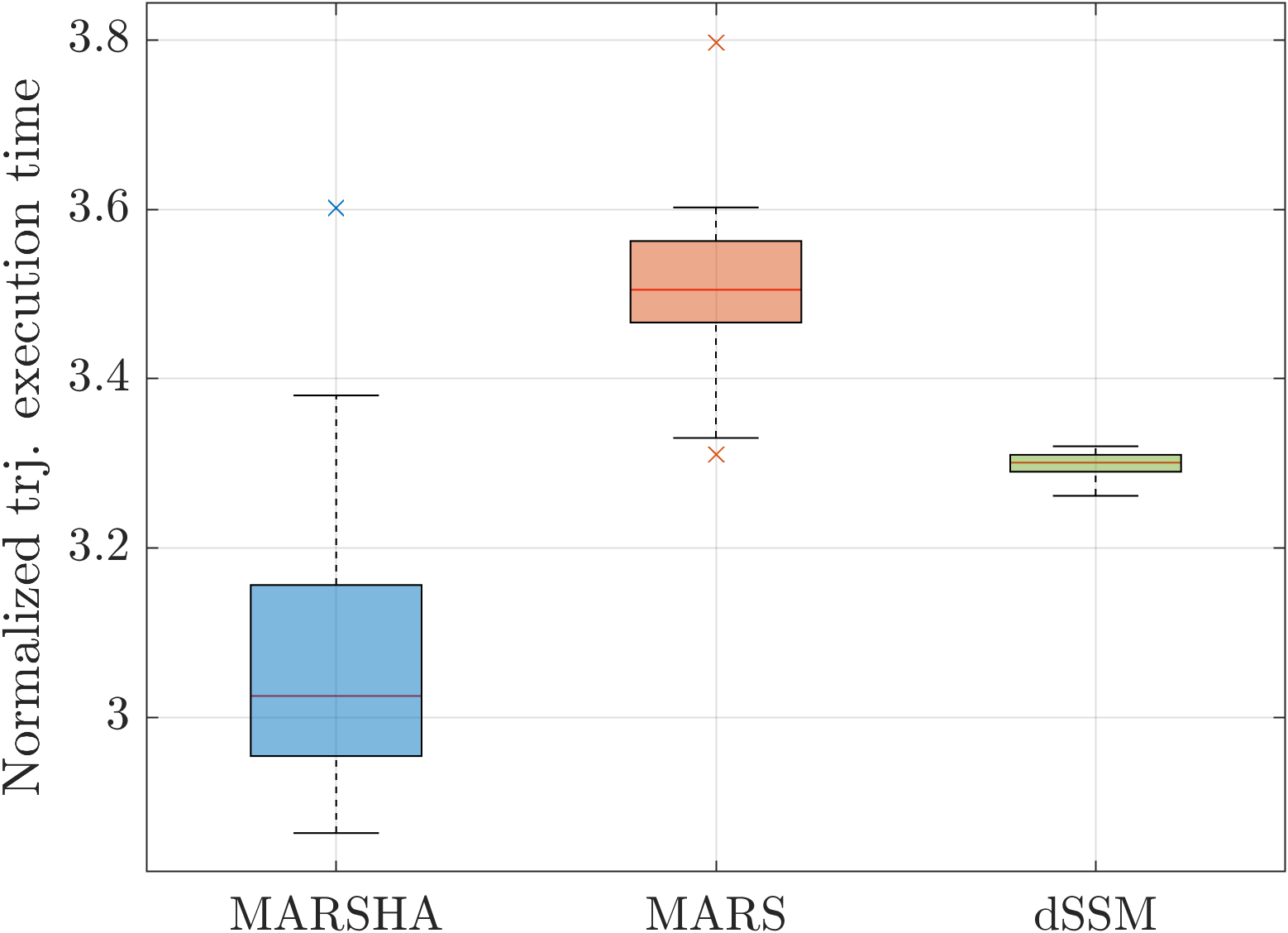}
    }
\caption{Comparison with reactive approaches. Maximum robot speed scaled to $30\%$ (top) and $60\%$ (bottom).}
\label{fig: speed_scaled}
\vspace{-0.5cm}
\end{figure}

\subsection{Comparison with Proactive Approaches}
\subsubsection{Simulation} \label{sec: proactive_test}

\begin{figure*}
\centering
    \captionsetup[subfloat]{labelformat=empty}
    \subfloat[]{%
        \includegraphics[height = 0.2\textwidth]{img/real_fast_exec_time_exp.png}
    }\qquad
    \subfloat[]{%
        \includegraphics[height = 0.2\textwidth]{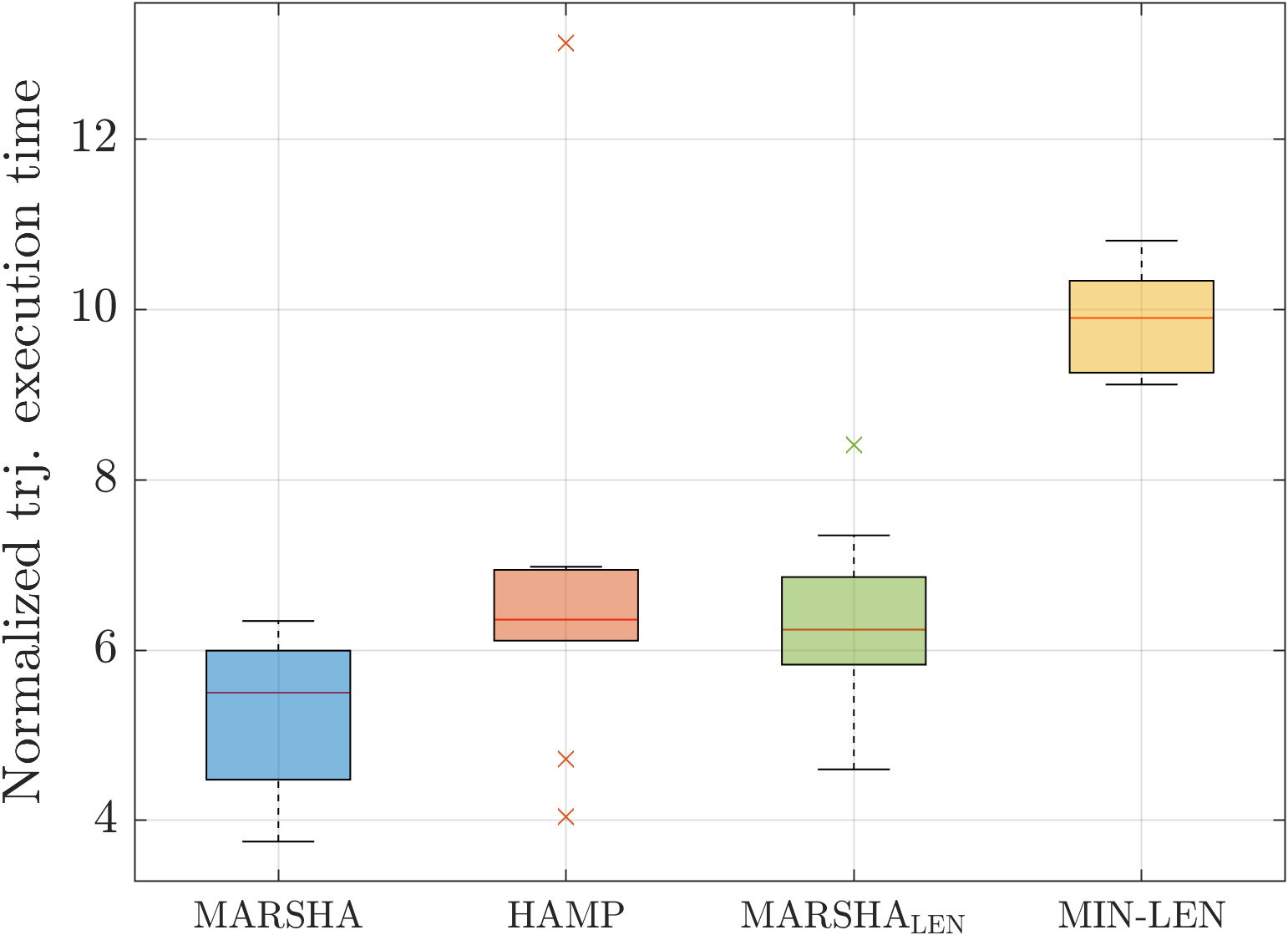}
    }\qquad\,
    \subfloat[ ]{%
        \includegraphics[trim={0.03cm 0cm 0cm 0.028cm},clip,height = 0.2\textwidth]{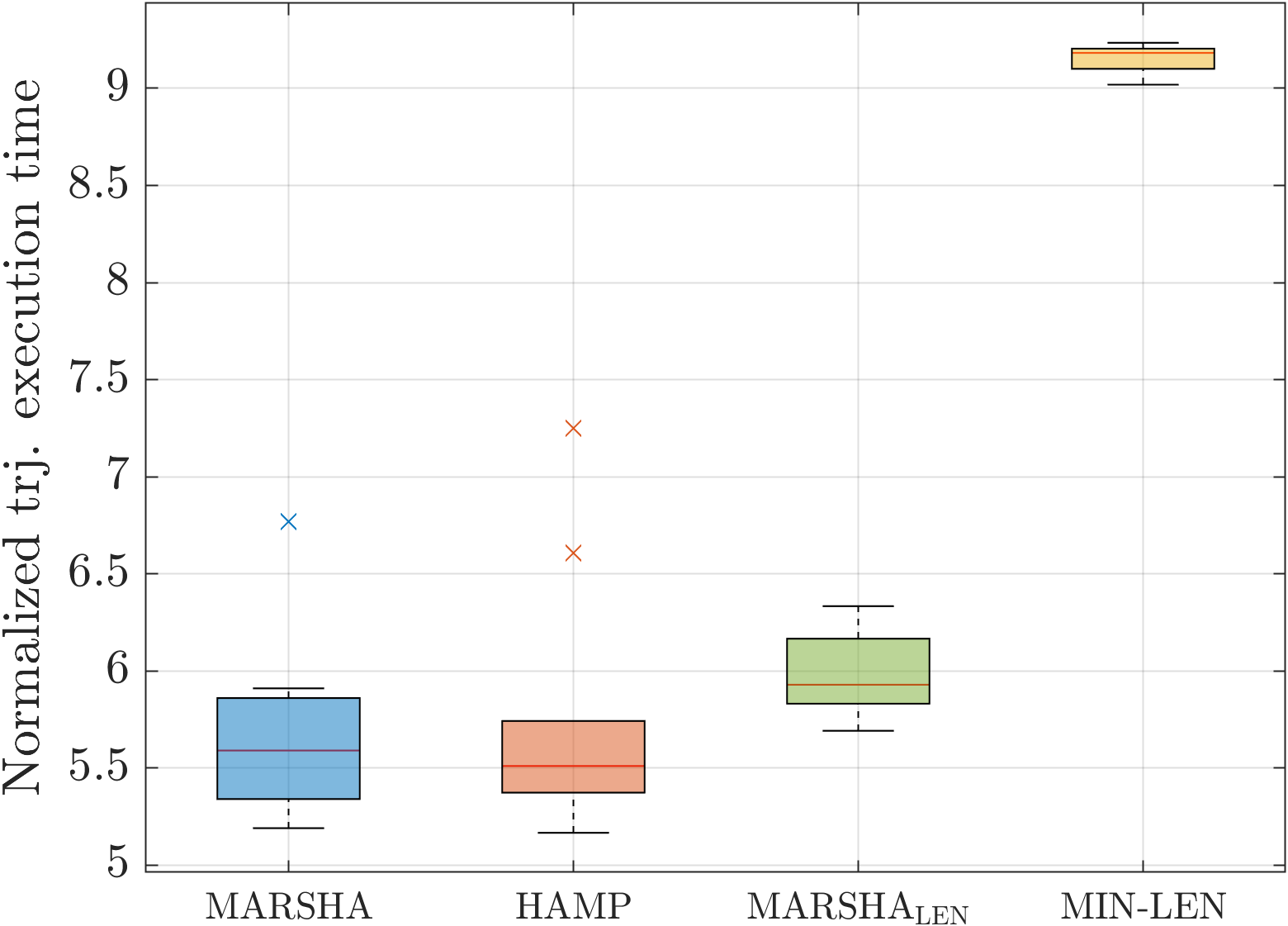}
    }
    \\
    \vspace{-0.5cm}
    \!\!
    \subfloat[(a)]{%
        \includegraphics[trim={0.04cm 0cm 0cm 0.0cm},clip,height = 0.2\textwidth]{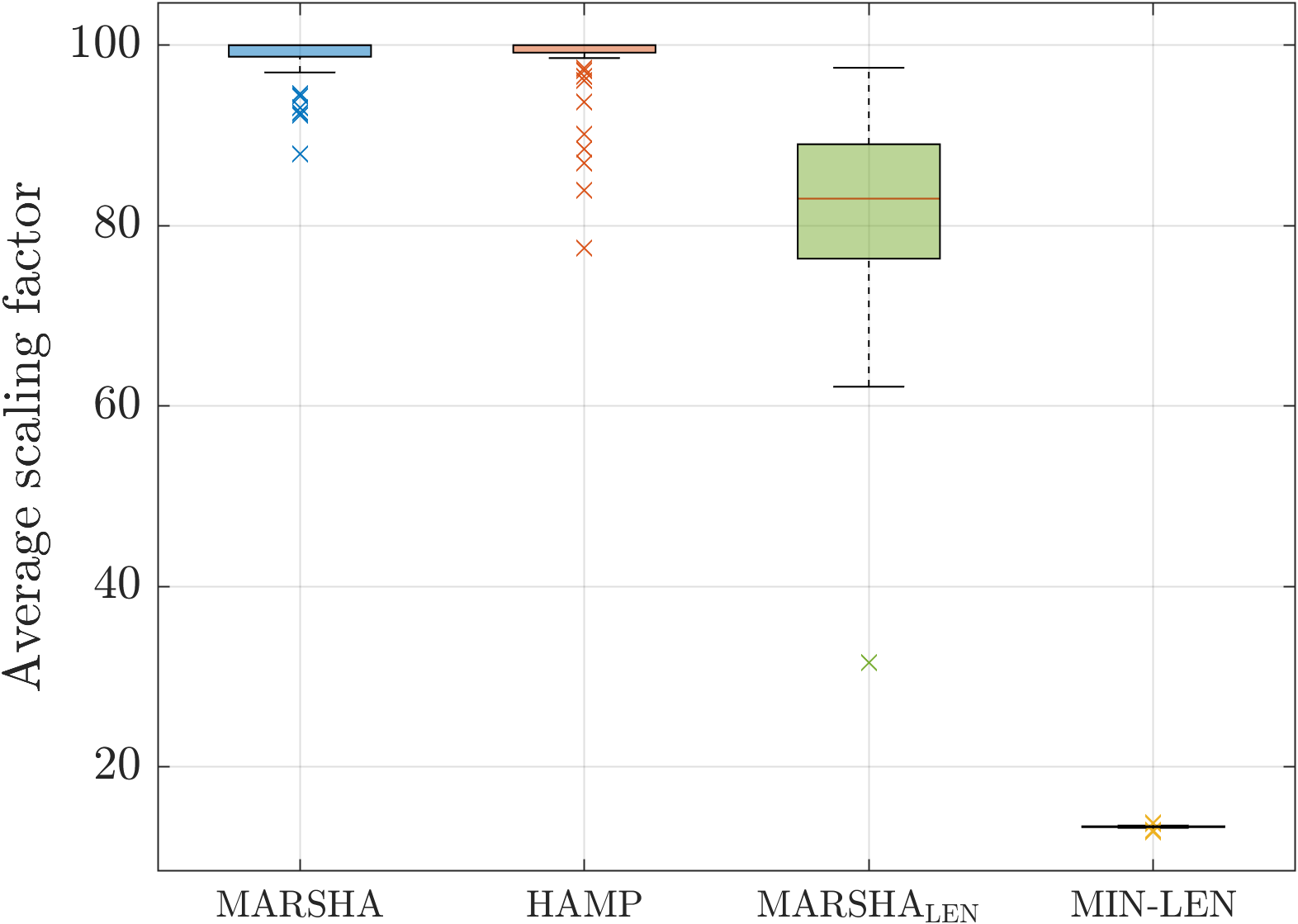}
    }\qquad
    \subfloat[(b)]{%
        \includegraphics[trim={0.04cm 0cm 0cm 0.02cm},clip,height = 0.2\textwidth]{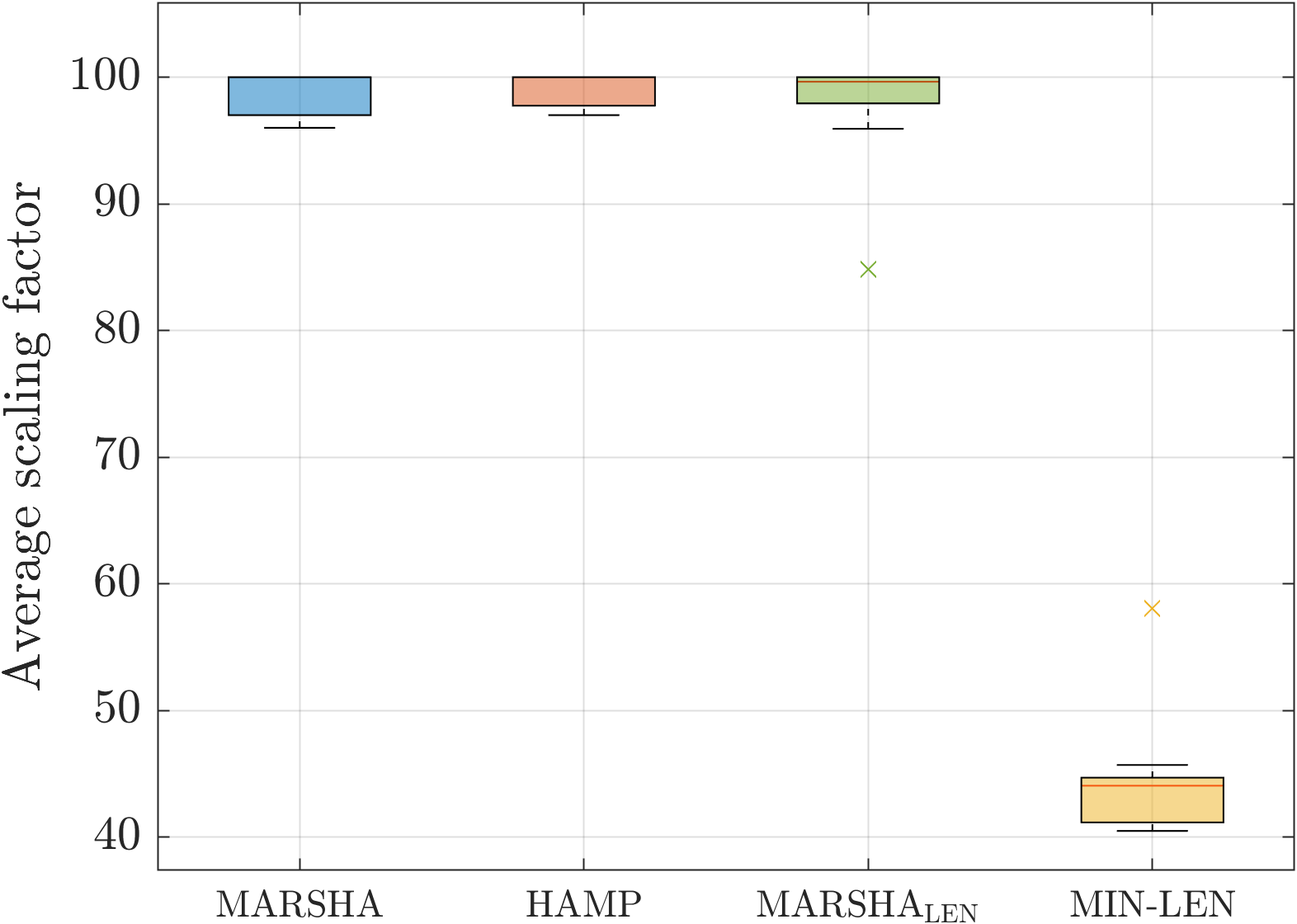}
    }\qquad
    \subfloat[(c)]{%
        \includegraphics[height = 0.2\textwidth]{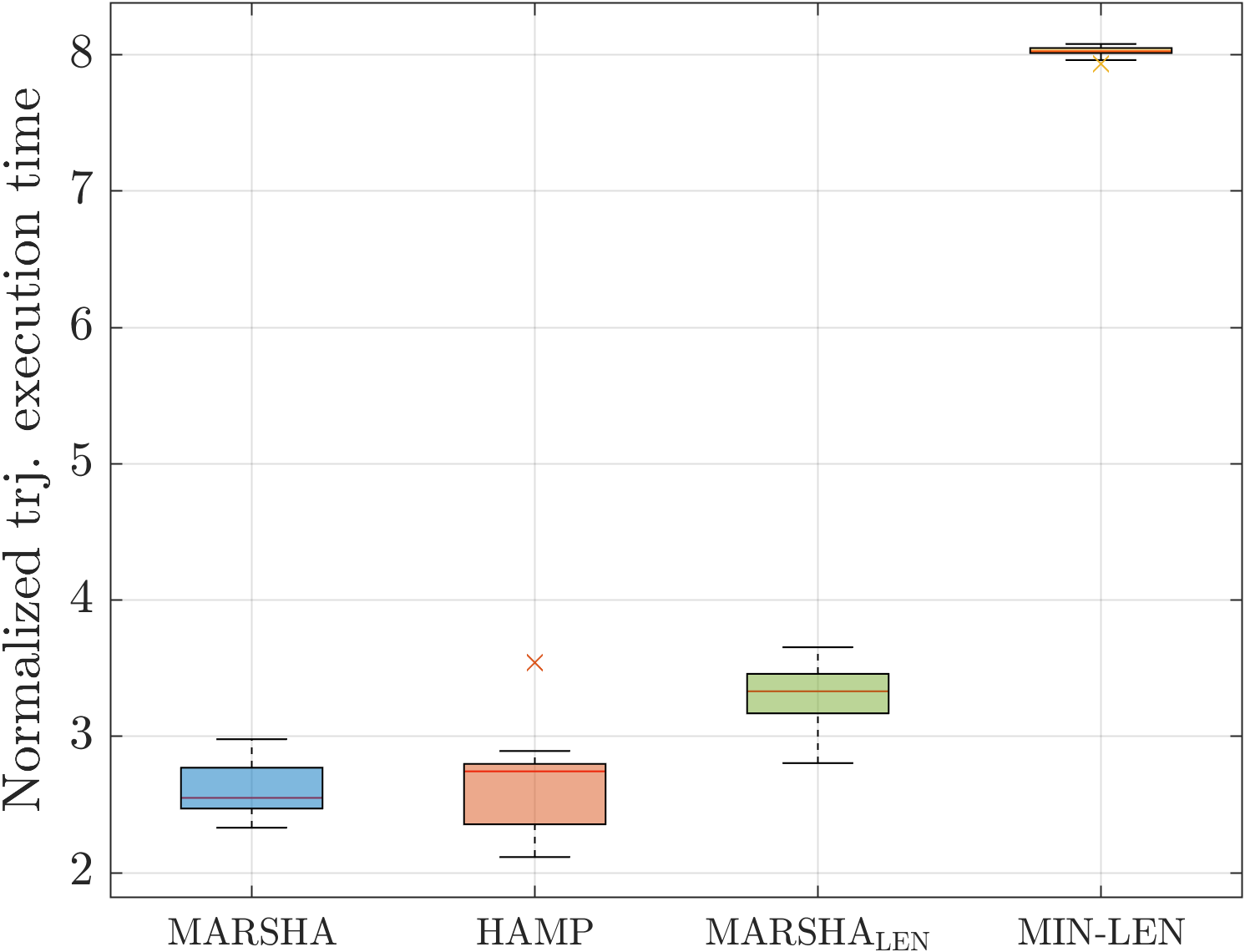}
    }
\caption{Test with proactive approaches. (a) simulations; (b) real experiments; (c) maximum robot speed scaled to $30\%$ (top) and $60\%$ (bottom).}
\label{fig: proactive_test}
\end{figure*}

In the tests, the robot moves from a starting position to a goal while the operator works in the shared space for $20$ seconds. When the initial path is calculated, the operator is already near the work table, allowing proactive approaches to factor in the operator's position during the planning phase. Each algorithm (HAMP \cite{HAMP}, MARSHA, and MIN-LEN \cite{RRT*}) was tested with $50$ simulations and $20$ trials in a real cell.
HAMP plans a path that minimizes execution time by considering the operator's position during the planning phase, while MIN-LEN finds the shortest path while avoiding collisions with the person. 
MARSHA was tested with initial paths calculated using both HAMP and MIN-LEN. Comparing these two methodologies helps us understand if the initial path calculation heavily influences MARSHA's performance. The results are illustrated in Fig. \ref{fig: proactive_test} (a) and (b). MARSHA\textsubscript{LEN} refers to MARSHA with the initial path computed using the MIN-LEN method.

The conducted tests yield some considerations.  When the initial path of MARSHA is safety-aware, HAMP and MARSHA exhibit comparable execution times. In some cases, MARSHA even outperforms HAMP by efficiently accommodating deviations caused by the operator. The fixed 20-second planning time of HAMP can lead to suboptimal solutions, whereas MARSHA's ability to improve the plan during execution results in time savings. As expected, MIN-LEN performs poorly due to its path's excessive proximity to the operator, which results in significant slowdowns of the shortest path.
However, an intriguing result emerges when the initial path of MARSHA is generated using MIN-LEN. Even without human-awareness on the initial path, MARSHA's performance remains relatively unaffected compared to using a human-aware path. MARSHA can provide performance similar to a human-aware planner using a simpler and faster generic initial path calculation that takes just a few seconds instead of tens of seconds. This outcome solidifies MARSHA's position as a hybrid control strategy, skillfully combining elements from proactive and reactive planners and showcasing unique characteristics from both categories.

\subsubsection{Real scenario} \label{sec: proactive_test_real}

We also observe a narrowing gap between MIN-LEN and the other algorithms in real experiments. This phenomenon is related to the constant scaling at $30\%$. When the operator is near the table, the robot halts, rendering the $30\%$ scaling effect insignificant during those $20$ seconds of stoppage. However, when the operator is far, allowing the robot to move freely, it follows the shortest path, and the constant scaling affects a short trajectory. 
Conversely, the other algorithms maintain a certain distance from the operator, leading to longer travels at a reduced speed of $30\%$. Consequently, the impact of the constant scaling becomes more pronounced for these algorithms. \rev{As in Section \ref{sec: reactive_test_real}, we repeated the simulations under constant speed scaling conditions of both $30\%$ and $60\%$, combined with the scaling imposed by the safety module. The results shown in Fig. \ref{fig: proactive_test} (c) demonstrate that lower robot velocities tend to reduce performance disparities between the algorithms.}

MARSHA displays superior adaptability to HAMP when the operator's behavior deviates from expectations, such as moving or walking away. 
HAMP's trajectory experiences slowdowns when the operator moves unexpectedly within the workspace, as it lacks online adaptation capabilities. 
Moreover, if the operator moves away from the designated area, HAMP continues to follow the same trajectory as if the operator were still in the shared workspace, resulting in a longer trajectory. In contrast, MARSHA quickly adjusts its plan, guiding the robot along the optimal trajectory to accommodate the operator’s movements within the workspace or when the operator exits. These results are visually demonstrated in the accompanying video, available at \cite{video}.

\subsection{Effect of the SSM Parameters on the Performance}  \label{sec: ssm_param}

\begin{table*}[t]
\caption{Values used for each parameter of  \eqref{eq: vmax SSM}.}
\label{tab:ssm_param}
\centering
\scalebox{0.9}{\begin{tabular}{|g|c|c|c|c|c|c|c|c|c|c|c|c|c|c|c|c|}
\hline
\rowcolor{Gray} set & 1 & 2 & 3 & 4 & 5 & 6 & 7 & 8 & 9 & 10 & 11& 12 & 13 & 14 & 15 & 16 \\

\hline
$C$ ($m$) & $0.10$ & $0.10$ & $0.10$ & $0.10$ & $0.10$ & $0.10$ & $0.10$ & $0.10$ & $0.30$ & $0.30$ & $0.30$ & $0.30$ & $0.30$ & $0.30$ & $0.30$ & $0.30$ \\
\hline
$T_r$ ($s$) & $0.15$ & $0.15$ & $0.15$ & $0.15$ & $ 0.30$& $ 0.30$& $ 0.30$& $ 0.30$& $0.15$ & $0.15$ & $0.15$ & $0.15$ & $ 0.30$& $ 0.30$& $ 0.30$& $ 0.30$ \\
\hline
$v_h$ ($m/s$) & $  0.00 $ & $  0.00 $ & $1.60 $ & $1.60 $ &  $   0.00$ & $   0.00$ & $ 1.60$ &   $ 1.60$ &   $  0.00 $ & $  0.00 $ & $1.60 $ &   $1.60 $ &   $   0.00$ & $   0.00$ & $ 1.60$ &   $ 1.60$   \\
\hline
$a_s$ ($m/s^2$) & $0.10$ & $2.50$ & $0.10$ & $2.50$ & $0.10$ &$2.50$ &$0.10$ & $2.50$ & $0.10$ & $2.50$ & $0.10$ & $2.50$ & $0.10$ &$2.50$ &$0.10$ & $2.50$  \\
\hline
\end{tabular}}
\end{table*}

\begin{figure*}[tbp]
    \centering
    \subfloat{%
    \includegraphics[width = 0.6\textwidth]{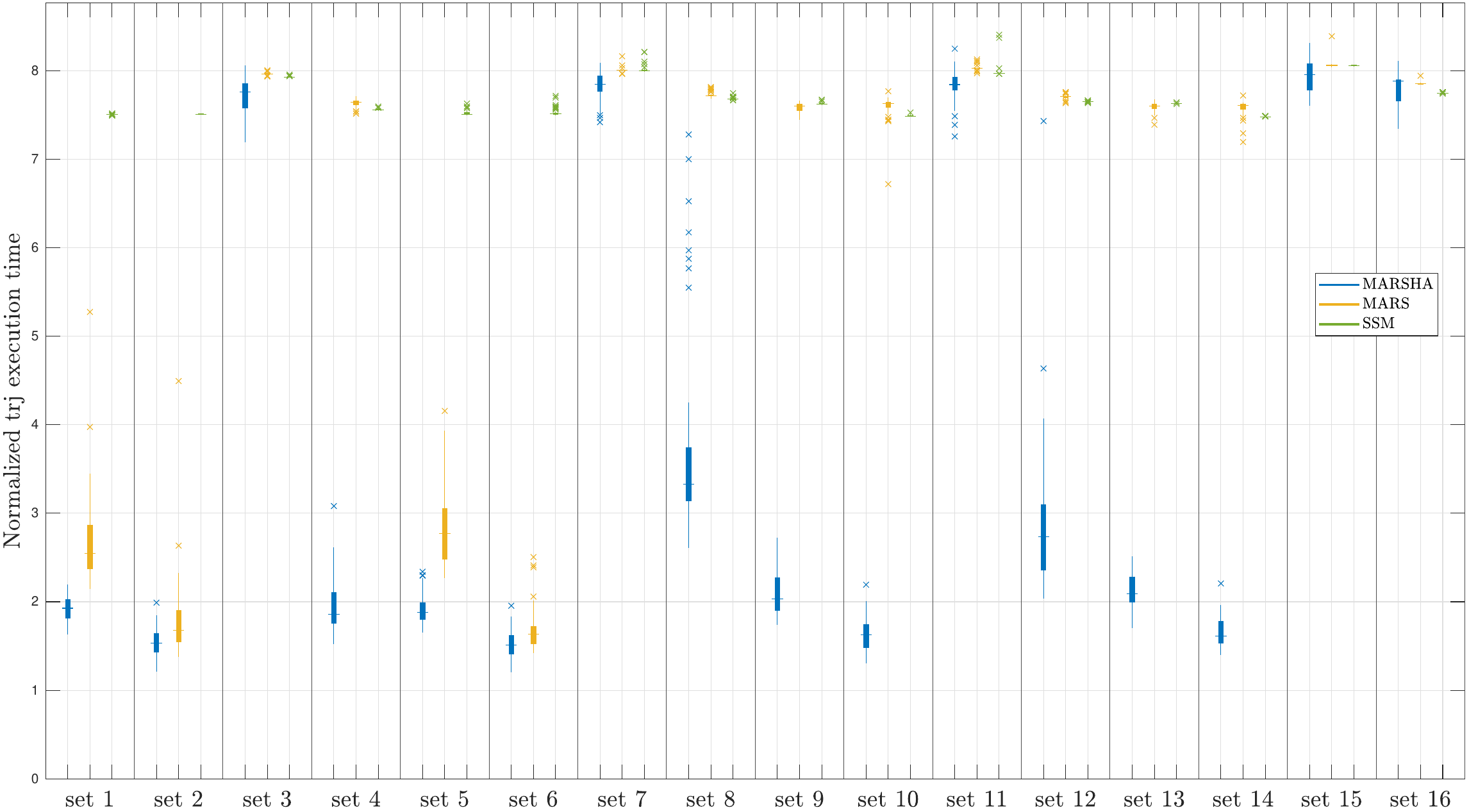}
    }
    \hfill
    \subfloat{%
    \includegraphics[width = 0.6\textwidth]{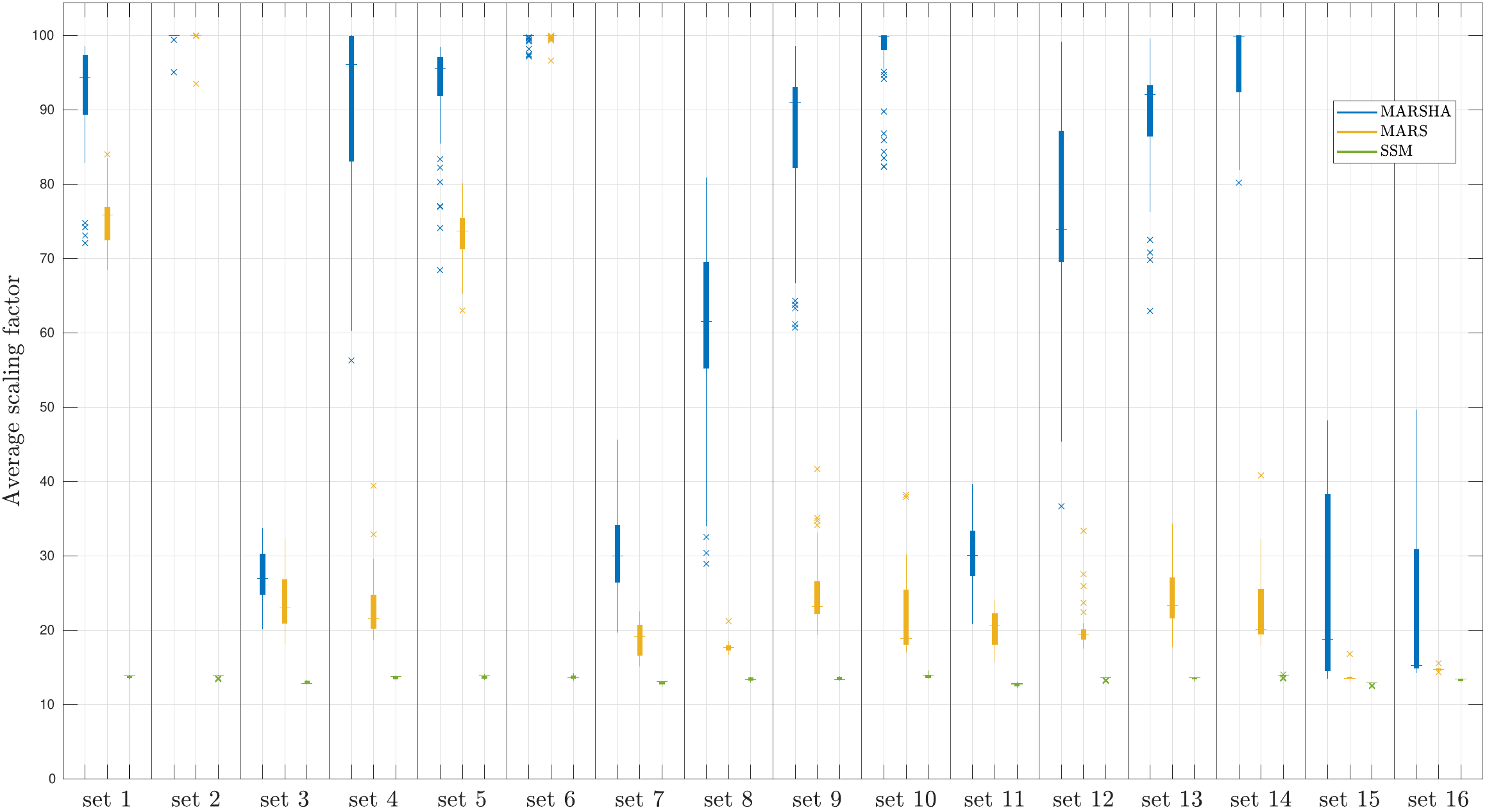}
    }
    \caption{\textit{Long} test with  the SSM parameter sets in Table \ref{tab:ssm_param}.}
\label{fig: ssm_param}
\end{figure*}
This section analyzes the impact of SSM parameters \eqref{eq: vmax SSM} on MARSHA's performance and how the performance gap with MARS/dSSM changes accordingly. 

Section \ref{sec: desing_exp} outlines the parameters used in the tests described in Sections \ref{sec: reactive_test} through \ref{sec: proactive_test_real}. By modifying these parameters, the degree to which the safety module intervenes to decelerate the robot varies. Certain parameter combinations prompt immediate and pronounced reactions from the safety module as soon as the operator enters the shared workspace, while other combinations allow for delayed intervention. The risk assessment must determine these parameters according to the system reaction times ($T_r$) and the robot's maximum Cartesian acceleration ($a_s$). 

Two values were chosen for each parameter, resulting in $16$ combinations. For each combination, the \textit{long} test was repeated $50$ times for each algorithm. Table \ref{tab:ssm_param} presents the selected parameter values, including those representing a more reactive cell and those simulating a less reactive cell. Two-speed values were considered for the human operator: $0~m/s$ and $1.6~m/s$, which aligns with the recommended value by ISO/TS 15066 when measuring the human walking speed is not feasible. Fig. \ref{fig: ssm_param} shows that varying sets of parameters have distinct effects on the performance of MARSHA.
MARSHA outperforms MARS/dSSM in most cases and demonstrates comparable performance in the worst-case scenarios.
With conservative implementations, MARSHA's behavior aligns with that of dSSM. The minimum allowed human-robot distance prevents MARSHA from generating solutions that do not trigger safety interventions. 
MARS achieves good performance levels when less conservative implementations allow for a smaller safety distance. In the remaining cases, MARSHA provides shorter execution times. Similar considerations can be drawn for PFL parameters.

\subsection{Summary of Results and Discussion}
Compared to reactive approaches (MARS and dSSM), MARSHA reduces the interventions of the safety module and reach the goal faster, especially during long operator interactions. The advantages are less significant for sporadic and brief interventions.
This result highlights the importance of the application's features in selecting the most effective planning approach.
Safety module configurations influence MARSHA's performance: conservative settings yield a behavior similar to dSSM, while with less conservative settings MARS already performs well. 
In the remaining scenarios, MARSHA achieves shorter execution times.
MARSHA outperformed offline planners (HAMP and MIN-LEN) by improving real-time solutions and adjusting to unforeseen operator movements. 
Through online optimization, MARSHA outperforms even offline planners when its initial path is not human-aware. 
This confirms the advantages of its proactive-reactive nature.

\section{Conclusions} \label{sec: conclusions}
We presented MARSHA, a motion replanning algorithm to enhance human-robot cooperation.
MARSHA dynamically estimates safety stops and slowdowns owing to human-robot proximity and continuously adjusts the robot's path in real time to minimize the estimated trajectory execution time. Unlike previous works, MARSHA blends reactivity with safety-aware proactivity. Compared to purely reactive or proactive approaches, MARSHA reduces the need for safety interventions, especially in long-term human-robot interactions. 
%
Its hybrid approach opens new possibilities for seamless human-robot cooperation across various application domains.
One potential limitation is its reliance on robust perception systems, which can be challenging in unstructured environments where obtaining reliable scene information may be difficult. Future work will extend the algorithm to explore additional interaction factors, such as legibility and ergonomics.

\bibliographystyle{IEEEtran}
\bibliography{IEEEabrv,reference}

\end{document}